\documentclass[DIV=calc, paper=a4, fontsize=11pt, twocolumn]{scrartcl}	 

\usepackage{lipsum} 
\usepackage[english]{babel} 
\usepackage[protrusion=true,expansion=true]{microtype} 
\usepackage{amsmath,amsfonts,amsthm} 
\usepackage[svgnames]{xcolor} 
\usepackage[hang, small,labelfont=bf,up,textfont=it,up]{caption} 
\usepackage{booktabs} 
\usepackage{fix-cm}	 

\usepackage{sectsty} 
\allsectionsfont{\usefont{OT1}{phv}{b}{n}} 

\usepackage{fancyhdr} 
\usepackage{lastpage} 

\usepackage{lettrine} 

\usepackage{titling} 

\newcommand{\HorRule}{\color{DarkGoldenrod} \rule{\linewidth}{1pt}} 

\pretitle{\vspace{-30pt} \begin{flushleft} \HorRule \fontsize{24}{28} \usefont{OT1}{phv}{b}{n} \color{DarkRed} \selectfont} 

\title{Appearance Descriptors for Person Re-identification: a Comprehensive Review}

\posttitle{\par\end{flushleft}\vskip 0.5em} 

\preauthor{\begin{flushleft}\large \lineskip 0.5em \usefont{OT1}{phv}{b}{sl} \color{DarkRed}} 

\author{Riccardo Satta\\}

\postauthor{\footnotesize \usefont{OT1}{phv}{m}{sl} \color{Black} 
DIEE - Dept. of Electrical and Electronic Engineering\\University of Cagliari, Italy\\ riccardo.satta@diee.unica.it

\par\end{flushleft}\HorRule} 

\date{July 2013} 

\usepackage{graphicx}
\usepackage{amsmath}

\begin{document}

\maketitle

\begin{abstract}
\textbf{Abstract.}
\textit{In video-surveillance, person re-identification is the task of recognising whether an individual has already been observed over a network of cameras. Typically, this is achieved by exploiting the clothing appearance, as classical biometric traits like the face are impractical in real-world video surveillance scenarios. Clothing appearance is represented by means of low-level \textit{local} and/or \textit{global} features of the image, usually extracted according to some part-based body model to treat different body parts (e.g. torso and legs) independently. 
This paper provides a comprehensive review of current approaches to build appearance descriptors for person re-identification. The most relevant techniques are described in detail, and categorised according to the body models and features used. The aim of this work is to provide a structured body of knowledge and a starting point for researchers willing to conduct novel investigations on this challenging topic.}
\end{abstract}

\section{Introduction}
\emph{Person re-identification}\cite{Doretto2011} consists of recognising an individual who has already been observed (hence the term \emph{re}-identification) over a network of video surveillance cameras. 
The topic is currently attracting much interest from researchers, due to the various possible applications such a technique can enable, e.g., \textit{off-line} retrieval of all the video-sequences where an individual of interest appears, whose image is given as query, or \textit{on-line} pedestrian tracking over multiple, possibly not-overlapping cameras (a task also known as \emph{re-acquisition} \cite{Gray2007evaluating}).

While several biometric traits can be in principle used to this aim, strong pose variations and unconstrained environments make the use of classical biometric traits like face difficult of impractical \cite{Doretto2011} with the typical sensors and setting of a surveillance network.
Therefore, researchers explored the use of cues that pose less constraints, at the expense of an intrinsically lower identification capability. Among them, clothing appearance is used in the most of re-identification methods, as a soft, session-based cue, that is relatively easy to extract, and exhibits uniqueness over a limited time span.
Various \emph{descriptors} of the clothing appearance have been proposed so far in the literature \cite{Doretto2011}. They are mostly designed heuristically, and are based on the extraction of various kinds of low-level \emph{local} and \emph{global} features from the images showing the individual\footnote{The term ``local features'' refers to localised characteristics of the image, e.g. the colour distribution around a certain salient point of the image; the term ``global features'', instead, refers to characteristics of the whole image, e.g. the overall colour distribution.}. Typically, they exploit a part-based body model, to take into account the non-rigid structure of the human body and treat the appearance of different body parts (e.g. torso and legs) independently. 

This paper provides an overview of existing methods used in literature for the task of person re-identification, with particular respect to the techniques used to build a descriptor of the body appearance. The presented review is mostly based on Chapter 2 of my thesis work \cite{Satta2013PHD}. The remainder of the paper is structured as follows. Sect.~\ref{sect:overview} first gives a simple formal statement to person re-identification. Then Sect.~\ref{sec:descriptors} reviews current approaches to construct appearance descriptors. The survey is conducted under two ``orthogonal'' viewpoints, namely the kind of body model and the kind of features used (Sect.~\ref{sec:bodymodelsSOA} and Sect.~\ref{sec:featuresSOA} respectively). Sect.~\ref{sec:combinationSOA} then focuses on the problem and current approaches of combining different feature sets 
While almost all existing methods use the clothing appearance as main cue to perform re-identification, it is worth to note that other approaches have been attempted in literature, for instance based on gait, or anthropometric measures captured through novel RGB-D sensors. These methods are briefly surveyed in Sect.~\ref{sec:othercuesSOA}. Finally, Sect.~\ref{sec:conclusions}, which concludes the paper.

\section{Problem overview}
\label{sect:overview}

Formally, person re-identification can be modelled as a \textit{recognition/matching} task, where a \emph{probe} individual is matched against a gallery of \emph{templates} (representing the individuals previously seen by the camera network).
Thus, the problem of re-identifying an individual represented by its descriptor $\mathbf P$ can be formulated as:
\begin{equation}
\label{eq:reidentification}
\mathbf T = \operatorname*{arg\,min}_{\mathbf T_i} D(\mathbf T_i, \mathbf Q) \ , \mathbf T_i \in \mathcal{T}
\end{equation}
where $\mathcal{T} = \{ \mathbf T_1, \ldots, \mathbf T_N \}$ is a gallery of $N$ template descriptors, and $D(\cdot,\cdot)$ is a proper distance metric.

In order to address the re-identification problem above, it is indeed fundamental, first, to answer the question of how to \textit{represent} persons using a descriptor. This is the topic of investigation of the rest of the paper.

\section{Appearance descriptors}
\label{sec:descriptors}

\begin{figure}[t]
\centering
   \includegraphics[width=0.90\linewidth]{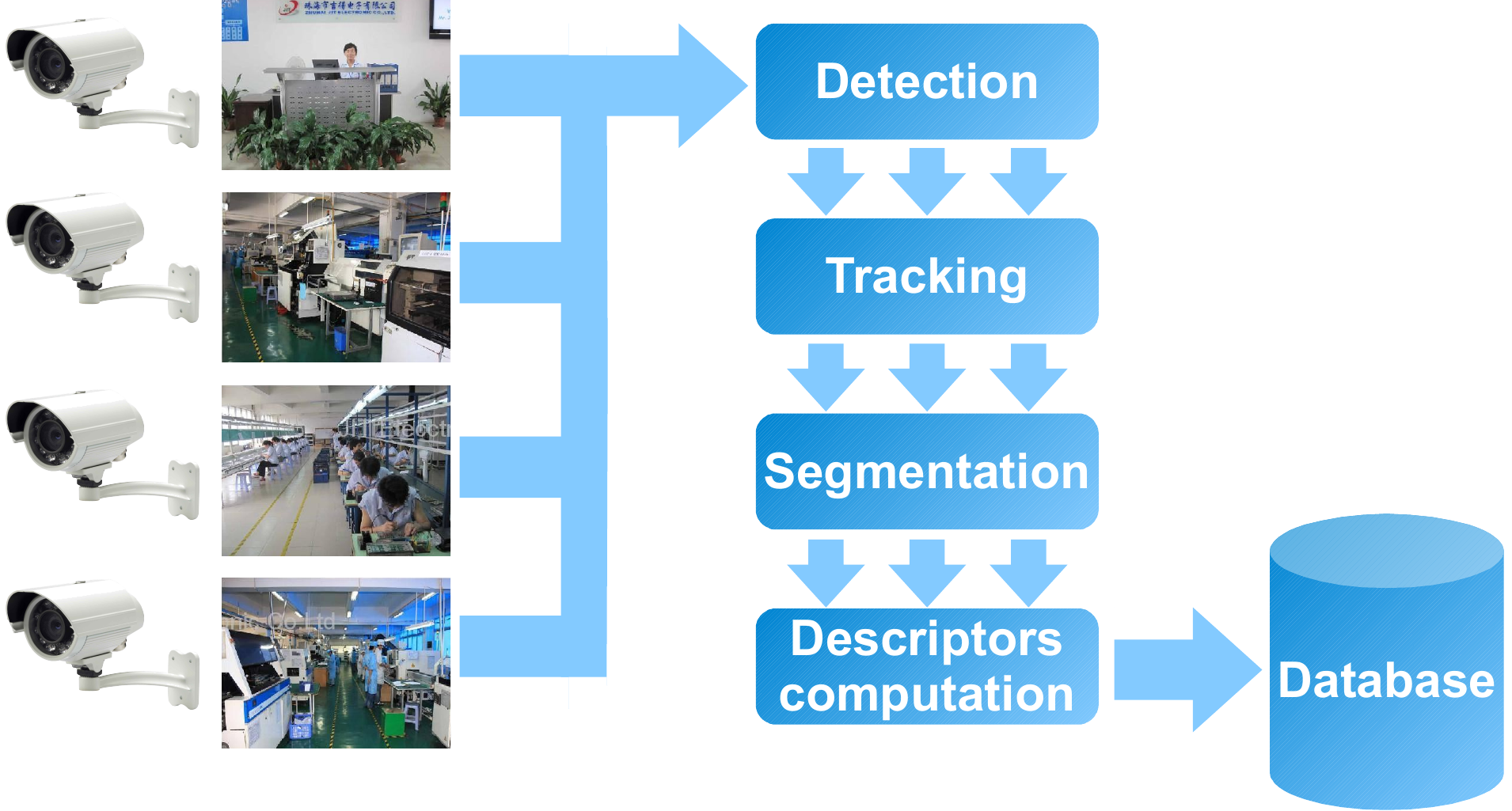}
   \caption{Descriptor construction pipeline.}
\label{fig:pipeline}
\end{figure}

\begin{figure}[t]
\centering
   \includegraphics[width=0.90\linewidth]{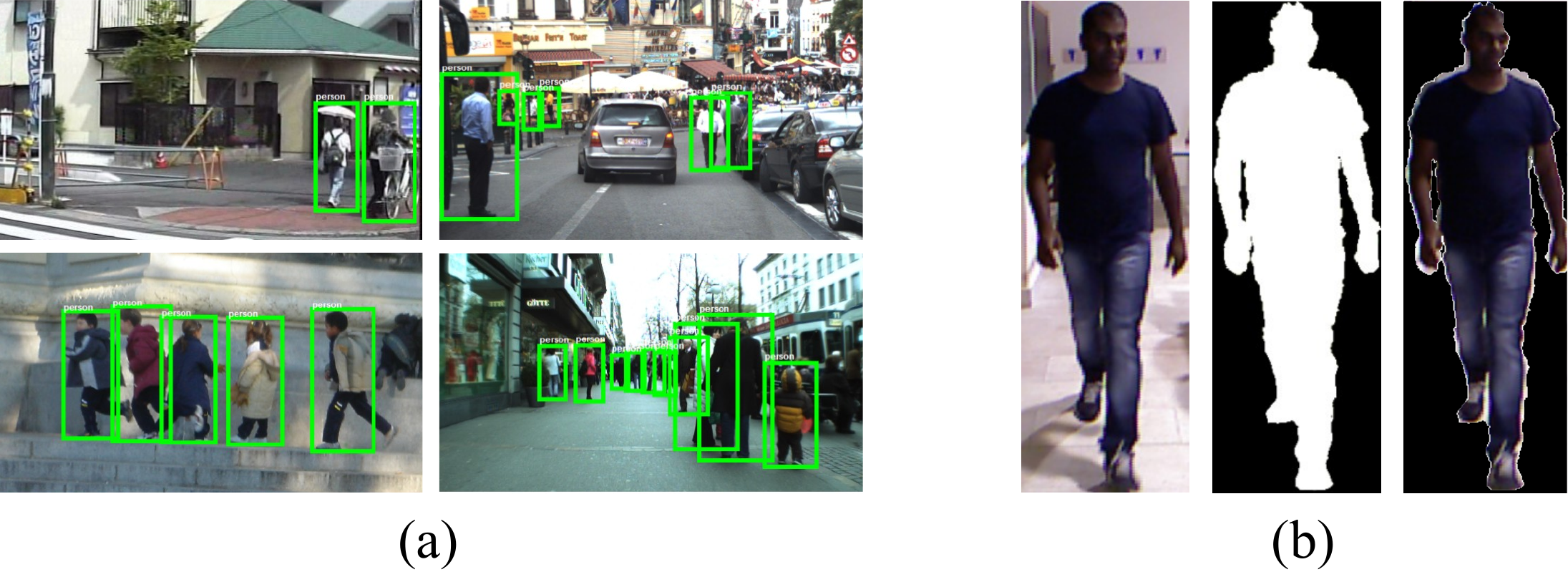}
   \caption{(a) Example outputs of a pedestrian detection algorithm in three frames taken from real-world video-surveillance footages; Detected blobs are in green. (b) Example of division of a blob into person and non-person pixels.}
\label{fig:pedDetection}
\end{figure}

The procedure of extracting appearance descriptors typically follow a standard pipeline (see Fig.~\ref{fig:pipeline} and Fig.~\ref{fig:pedDetection}):
\begin{enumerate}
\item the person is \emph{detected} and \emph{tracked} by suitable algorithms;
\item the pixels belonging to the person are separated from the background (\emph{foreground extraction} or \emph{segmentation}) in each frame of the video-sequence;
\item a descriptor is built from the resulting silhouettes (one for each frame), using \emph{local} or \emph{global} features, possibly after different body parts are detected through a body model, in order to take the into account the non-rigid nature of the body;
\end{enumerate}
Descriptors of Step 3 are finally stored in a data base for subsequent searches.


Step 1 requires i) a method to detect people in a given video frame \cite{Dollar2012} (i.e., to recognise the image regions, or \emph{blobs}, that contain a person), and ii) a data association algorithm that track people found by the detector \cite{Isard2001,Niu2003,Yilmaz2006} (i.e., to associate blobs in subsequent frames to the same person). These two steps may also be carried out together, and reinforce one another \cite{Andriluka2008tracking}.
Step 2 is usually carried out using an adaptive model of the background \cite{Elgammal2000}.

\begin{figure}[t]
\centering
   \includegraphics[width=0.99\linewidth]{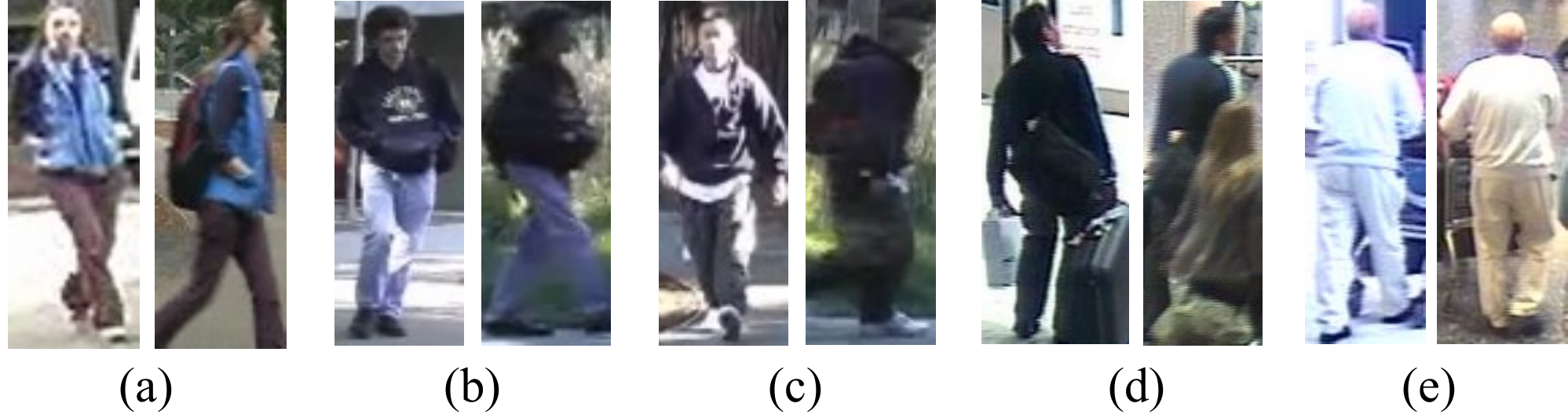}
   \caption{Pairs of images showing the same person from different cameras, taken from two common benchmark data sets, VIPeR \cite{Gray2007evaluating} and i-LIDS \cite{Zheng2009ilids}. Notice pose variations (a)(b)(c), partial occlusions (d), illumination changes (a)(b)(c), and different colour responses (e).}
\label{fig:exampleIssues}
\end{figure}

Many challenging issues can affect some or all the three steps above. Among them we cite (see Fig.~\ref{fig:exampleIssues}):
\begin{itemize}
\item \textbf{Pose and viewpoint variations.} The relative pose of a person with respect to the cameras of the network varies depending on the walking path of that person, and of the viewpoint of the camera. This may cause consistent variations of the person appearance.
\item \textbf{Partial occlusions.} Parts of a person may be not visible to the camera due to occlusions caused by objects, clothing accessories or other people. This may cause the segmentation algorithm to fail in separating one person from the rest of the scene; consequently, descriptors may be built from images partially corrupted by the source of the occlusion.
\item \textbf{Illumination changes.} Illumination conditions may differ in different cameras, and in the same camera in different periods of time due to changing environmental conditions. This may result in appearance changes over different cameras and during time.
\item \textbf{Changes in colour response.} Different cameras may have a different colour response, that may affect person appearance as well.
\end{itemize}

The vast majority of methods assumes that the steps of detection, tracking and segmentation have been already accomplished using any of the algorithms available in literature, and concentrate on the task of constructing descriptors. The interested reader is referred to \cite{Bouwmans2008} and \cite{Dollar2012} for a comprehensive survey of pedestrian detection and foreground segmentation algorithms. 
This paper concentrates on Step 3, namely, how to construct discriminant and robust appearance descriptors to match persons in different views.

As stated in the introductory Section, appearance descriptors usually follow a part-based body model: the body is at first subdivided in \emph{parts}. Then, body parts are described via global features or bags (i.e., unordered sets) of local features.
Therefore, it is convenient to split the survey of current appearance descriptors in two parts, first reviewing body part subdivision models (Sect.\ref{sec:bodymodelsSOA}), then focusing on appearance features (Sect.\ref{sec:featuresSOA}). Combining different kind of features may help in attaining a better performance; Sect. \ref{sec:combinationSOA} provides a closer insight on typical approaches for feature combination in appearance descriptors.

\subsection{Part-based body models}
\label{sec:bodymodelsSOA}
The human body is not a rigid object. Instead, it has a complex kinematics, and can be better described using a part-based model, possibly where relative positions of parts are not fixed a-priori but are inferred from the image. Furthermore, discontinuities of the clothing appearance usually follow the body structure (e.g., the clothing appearances of the upper and lower body usually differ). 
Many existing appearance descriptors, therefore, exploit some part-based human body model to segment the silhouette into different parts.
Some other descriptors (e.g., \cite{Ayedi2012,Bazzani2010,Bouma2012,Dangelo2011,Gray2008ELF,Hamdoun2008,Hamdoun2008b,Hirzer2012,Javed2005,Kai2011,Ma2012,
Manjunath2001,Piccardi2005,Truong2009,TruongCong2010,zhao2013}) consider the body as a whole instead.
Part-based body models used in existing appearance descriptors can roughly be divided into three categories:
\begin{itemize}
\item \emph{fixed models}, in which size and relative position of body parts are defined a-priori;
\item \emph{adaptive models}, that try to fit a predefined part subdivision model to the image of the person;
\item \emph{learned models}, that previously learn the model constraints (e.g., relative parts disposition) from a labelled training set of images of individuals.
\end{itemize}
In the rest of this Section, part-based body models belonging to the three categories above are reviewed and compared.

\subsubsection{Fixed part models}
Probably the simplest kind of part subdivision is a fixed one, in which the sizes and positions of body parts are chosen a-priori. An example of this approach can be found in \cite{Liu2012,ProsserReIdBMVC2010,Wei-Shi2011}, where the body is subdivided into six horizontal stripes of equal size, that roughly capture the head, upper and lower torso and upper and lower legs. Similarly, in \cite{Avraham2012} the silhouette is subdivided in five equal-sized stripes.
An even simpler fixed part subdivision is used in \cite{Khan2010}. Three horizontal stripes of respectively 16\%, 29\% and 55\% of the total blob height roughly locate head, torso and legs, then the first strip is discarded as the head typically consists of few pixels and is not informative for the clothing appearance.

\subsubsection{Adaptive part models}
\label{subsec:adaptivePartModelsSOA}
Other body models are \emph{adaptive}, in the sense that they try to fit a predefined part subdivision model to the image of the individual.
In one of the descriptors proposed in \cite{Bak2010-2}, the MPEG-7 Dominant Colour Descriptor (DCD) \cite{Yang2008} is used to dynamically separate the body into two parts, upper and lower body, looking for discontinuities in dominant colours (the same DCD is also used as feature set to describe each body part, see Sect.~\ref{sec:featuresSOA}).
The approach of \cite{Farenzena2010} extends the basic idea of exploiting appearance anti-symmetries of \cite{Bak2010-2}. It dynamically finds three body areas, namely the head, torso, and legs, exploiting symmetry and anti-symmetry properties of silhouette and appearance.
To this aim, two operators are defined.
The first measures is called \emph{chromatic bilateral operator}. It measures the appearance anti-symmetry of a certain image region with respect to a given horizontal axis, and is defined as
\begin{equation}
\label{eq:cboperator}
 C(y,\delta) = \sum_{\substack{B_{[y-\delta,y+\delta]}}} d^2\big(p_i,\hat{p}_i\big),
\end{equation}
where $d(\cdot,\cdot)$ is the Euclidean distance, evaluated between pixels represented in the HSV colour space $p_i$ and $\hat{p}_i$ located symmetrically with respect to an horizontal axis placed at height $y$ of the person image. This distance is summed up over the person pixels lying in the horizontal strip $B_{[y-\delta,y+\delta]}$ centred in $y$ and of height $2\delta$.

\begin{figure}[t]
\centering
   \includegraphics[width=1.00\linewidth]{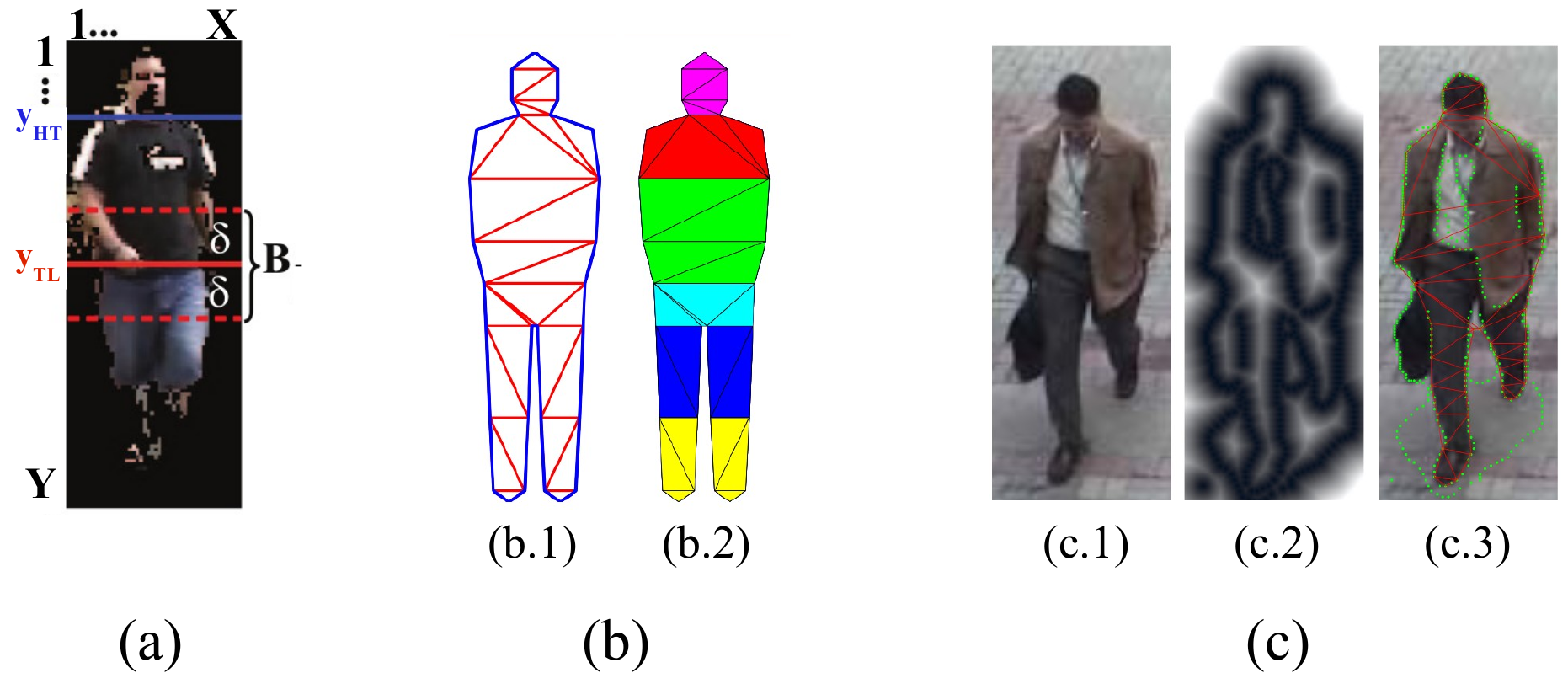}
   \caption{(a) Symmetry-driven subdivision in three parts \cite{Farenzena2010}. The blob of size $Y \times X$ pixels containing the person is divided according to two horizontal axes, $y_{HT}$ and $y_{TL}$, found by minimising a proper combination of the operators defined in Eqs.~\eqref{eq:cboperator}-\eqref{eq:scoperator}. (b) Decomposable body model used in \cite{Gheissari2006}: (b.1) the decomposable triangulated graph model; (b.2) Partitioning of the person according to the decomposable model. (c) An example of fitting the decomposable triangulated model of \cite{Gheissari2006} to an individual: (c.1) an image of an individual; (c.2) edges detected through the the Canny's algorithm \cite{Canny1986}; (c.3) result of fitting the model to the edges (in red). All figures are taken from \cite{Farenzena2010} and \cite{Gheissari2006}.}
\label{fig:partSubdivisions}
\end{figure}

The second is called \emph{spatial covering operator} and measures the difference of the silhouette areas of two regions:
\begin{equation}
\label{eq:scoperator}
 S(y,\delta) = \frac{1}{W\delta}\Big|A\big(B_{[y-\delta,y]}\big) - A\big(B_{[y,y+\delta]}\big)\Big|,
 \end{equation}
where $W$ is the width of the blob, and $A\big(B_{[y-\delta,y]}\big)$ and $A\big(B_{[y,y+\delta]}\big)$, denote the number of person pixels respectively of the strip of vertical extension $[y-\delta,y]$ and $[y,y+\delta]$.
These operators are combined to find two axes, $y_{HT}$ and $y_{TL}$, that respectively separate head and torso, and torso and legs. These axes are defined as
\begin{equation}
\label{eq:itl}
y_{TL} = \operatorname*{arg\,min}_{y}\big(1-C(y,\delta) + S(y,\delta)\big),
\end{equation}
\begin{equation}
\label{eq:iht}
y_{HT} = \operatorname*{arg\,min}_{y}\big(-S(y,\delta)\big).
\end{equation}
The parameter $\delta$ is set to a value of $\delta=Y/4$ where $Y$ is the blob height in pixels. The values $y_{HT}$ and $y_{TL}$ isolate three regions approximately corresponding to head, body and legs (Fig.~\ref{fig:partSubdivisions}-a). The head part is discarded as it carries very low informative content.
As claimed by the authors, this strategy is able to locate body parts which are dependent on the visual and positional information of the clothes, robust to pose, viewpoint variations, and low resolution. After \cite{Farenzena2010}, the same part-based model has been used in various other works \cite{Bazzani2012,Martinel2012EL,Martinel2012,Satta2011MCD,Satta2012MCD,Satta2011MCM,Wu2011}.

A deformable model that is fitted to each individual to find six body regions is used one of the methods in \cite{Gheissari2006}, based on decomposable triangulated graphs \cite{Amit1996}. A triangulated graph is a collection of cliques of size three, that has a perfect elimination order for their vertices, i.e., there exists an elimination order for all vertices such that (i) each eliminated vertex belongs only to one triangle, and (ii) a new decomposable triangulated graph results from eliminating the vertex.

The model is fit to the image of a person using the following strategy. Let the model be a decomposable triangulated graph T with $n$ triangles $T_i,i=1,\ldots,n$. The goal is to find a function $g$ that maps the model to the image domain, such that the consistency of the model with salient image features is maximised, and deformations of the underlying model are minimised.
The function $g$ must be a piecewise affine map \cite{Felzenszwalb2005}, i.e the deformation of each triangle $g_i(T_i)$ must be an affine transformation. 
The problem becomes to minimise an energy functional $E(g, I)$ that can be written as a sum of costs:
\begin{equation}
	E(g,I) = \sum_{i} E_i(g_i,I) = \sum_{i} \Big(E^{data}_i(g_i,I) + E^{shape}_i(g_i)\Big),
\end{equation}
where the $I$ represents the image features.
The terms $E^{shape}_i(g_i)$ take into account the cost for shape distortion of the $i$-th triangle, while $E^{data}_i(g_i,I)$ attracts the model to salient image features, which are found using an edge detector (Canny's algorithm \cite{Canny1986}).
As shown in \cite{Amit1996}, a model based on decomposable triangulated graphs can be efficiently optimised using dynamic programming. Once the model has been fitted with regard to the image, the individual is partitioned into six salient body parts, shown Fig.~\ref{fig:partSubdivisions}-b with different colours. An example of application to a real pedestrian image is shown in Fig.~\ref{fig:partSubdivisions}-c.

\subsubsection{Learned part models}
\label{subsec:learnedPartModelsSOA}
\begin{figure}[t]
\centering
   \includegraphics[width=0.99\linewidth]{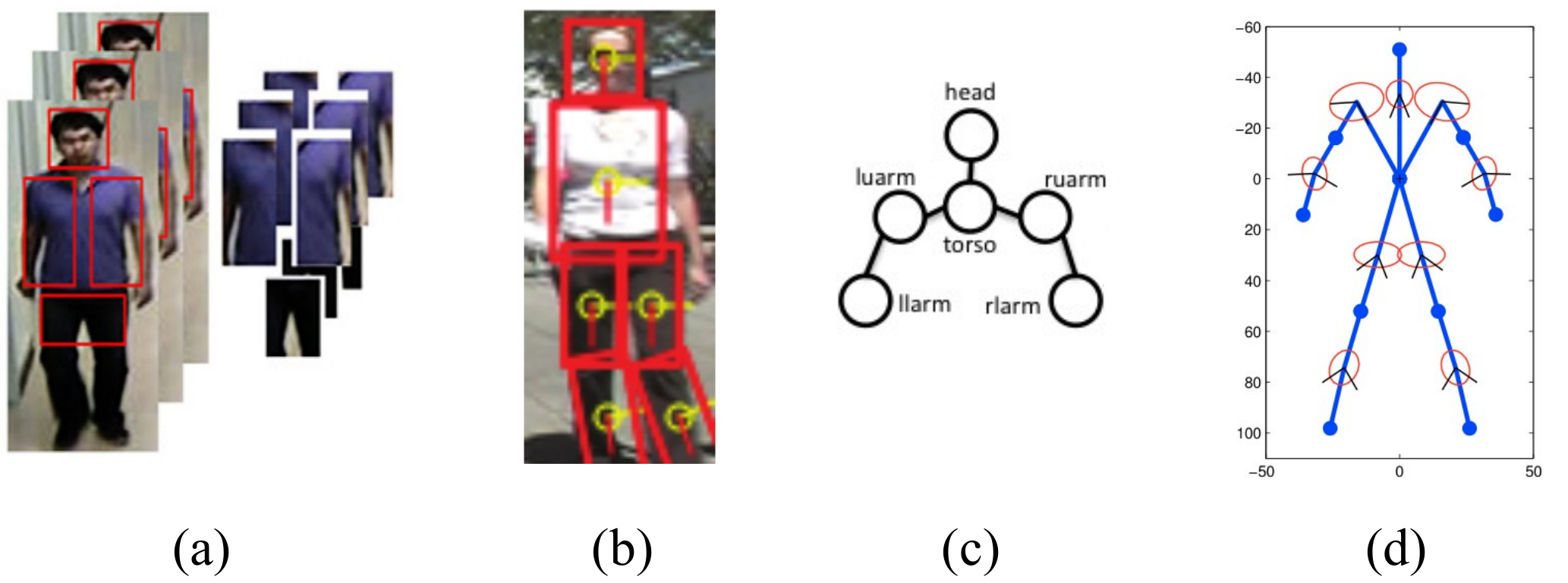}
   \caption{(a) Sample output of the articulated body model used in \cite{Bedagkar2012,Bedagkar2011}. (b) Sample output of the Pictorial Structure model used in \cite{Cheng2011}. (c) Sample Pictorial Structure of the upper body part, with the torso part as root node. (d) Kinematic prior learned on the dataset from \cite{Ramanan2006}. The mean part position is shown in blue dots; the covariance of the part relations in the transformed space is shown using red ellipses. Figures taken from \cite{Bedagkar2012} and \cite{Andriluka2009Pictorials}.}
\label{fig:articulatedModels}
\end{figure}

More recently, some methods that rely on previously trained body part detectors and articulated body models have been proposed. Part detectors are statistical classifiers that learn a model of a certain body part (e.g., an arm) from a given training set of images of people where body parts are manually located and labelled. Typically, these detectors exploit features related to the edges contained on the image.
An approach of this kind has been used in \cite{Bedagkar2012,Bedagkar2011} based on the work of Felzenszwalb et al. \cite{Felzenszwalb08}. The overall body model is made up of different part models; each one, in turn, consists of a \emph{spatial model} and of a \emph{part filter}. The spatial model defines a set of allowed placements for a part with respect to the bounding box containing the person, and a deformation cost for each placement. To learn a model, a generalisation of Support Vector Machines (SVM) \cite{Burges1998} called latent variable SVM (LSVM) is used.
In \cite{Bedagkar2012,Bedagkar2011}, such model is used to detect four different body parts, namely head, left torso, right torso and the upper legs (see Fig.~\ref{fig:articulatedModels}-a).

An articulated body model based on Pictorial Structures (PS) was proposed in \cite{Andriluka2009Pictorials} and later exploited in \cite{Cheng2011} for the task of re-identification. In \cite{Cheng2011}, six parts are considered (chest, head, thighs and legs, see Fig.~\ref{fig:articulatedModels}-b), while the original PS model is also able to detect and locate upper and lower arms.

A PS model for an object \cite{Felzenszwalb2005PS} is a collection of parts with connections between certain pairs of parts (an example is provided in Fig.~\ref{fig:articulatedModels}-c).
The approach of \cite{Andriluka2009Pictorials} uses a PS of the human body that is made up of a set of $N$ parts, and a set of generic part detectors based on descriptors of the shape. The model and the body part detectors are trained on a training set of images of people.

Let $L=\big\{\mathbf{l}_0,\ldots,\mathbf{l}_{N-1}\big\}$ be the set of configurations of each body part. Each $\mathbf{l}_i$ is the \emph{state} of the $i$-th body part $\mathbf{l}_i = \big(x_i,y_i,\theta_i,s_i\big)$, where $x_i$ and $y_i$ are the image coordinates of the part centre, $\theta_i$ is the absolute part orientation, and $s_i$ is the part scale, relative to the size of the part in the training set.
Given the image evidence $D$, the problem is to maximise the a-posteriori probability (\emph{posterior}) $p(L|D)$ that the part configuration $L$ is correct. The posterior is proportional to
\begin{equation}
\label{eq:PSposterior}
p(L|D) \propto p(D|L)p(L)
\end{equation}
according to Bayes' theorem \cite{DudaHart2001}. The term $p(D|L)$ is the likelihood of the image evidence given
a particular body part configuration, while $p(L)$ corresponds to a kinematic tree prior. Both are learned from a training set, as follows.

\textbf{Kinematic three prior.} The prior $p(L)$ encodes the kinematic constraints, i.e. the constraints on the relative parts disposition. The body structure is mapped on a directed acyclic graph, so that $p(L)$ can be factorised as
\begin{equation}
\label{eq:PSprior}
p(L) = p(\mathbf{l}_0) \prod_{\substack{(i,j)\in E}} p\left(\mathbf{l}_i|\mathbf{l}_j\right)
\end{equation}
where $E$ denotes the set of all directed edges in the kinematic tree, and $\mathbf{l}_0$ is the root node, that in \cite{Andriluka2009Pictorials} is chosen to be the torso body part.

The prior for the root part configuration $p(\mathbf{l}_0)$ is assumed to be uniform.
To model part relations $p(\mathbf{l}_i|\mathbf{l}_j)$, a transformed space is used, where such relations can be modelled as Gaussian \cite{Felzenszwalb2005PS}. More specifically, the part configuration $\mathbf{l}_i = \big(x_i,y_i,\theta_i,s_i\big)$ is transformed into the coordinate system of the joint between the two parts $i$ and $j$ using the transformation:
\begin{equation}
T_{ji}(\mathbf{l}_i )=
 \begin{pmatrix}
  x_i + s_i d^{ji}_x cos \theta_i + s_i d^{ji}_y sin \theta_i\\
  y_i + s_i d^{ji}_x sin \theta_i + s_i d^{ji}_y cos \theta_i\\
  \theta_i + \bar{\theta}_{ji}\\
  s_i
 \end{pmatrix}
\end{equation}
where $d^{ji} = \big(d^{ji}_x, d^{ji}_y\big)^T$ is the mean relative position of the joint between the two parts $i$ and $j$, in the coordinate system of part $i$, and $\bar{\theta}_{ji}$ is the relative angle between the two parts. Then, part relations are modelled as Gaussian in the transformed space:
\begin{equation}
p\left(\mathbf{l}_i|\mathbf{l}_j\right) = \mathcal{N}\left(T_{ji}(\mathbf{l}_i )|T_{ij}(\mathbf{l}_j),\Sigma^{ji}\right)
\end{equation}
where $d^{ji}$ and $\Sigma^{ji}$ can be learned via maximum likelihood estimation \cite{DudaHart2001} from a labelled training set of images of people. It is worth noting that the body parts are only loosely attached to the joints (also called a \emph{loose-limbed model} \cite{Sigal06}), which helps increasing the robustness of the pose estimation.
Fig.~\ref{fig:articulatedModels}-d shows the priors learned from the multiple views and multiple poses people data set of \cite{Ramanan2006}, a common benchmark corpus for body pose estimation algorithms.

\textbf{Likelihood of the image evidence.} 
To estimate the likelihood $p(D|L)$, the methods relies on a different appearance model for each body. Each appearance model will result in a part evidence map $\mathbf{d}_i$ that reports the evidence for the $i$-th part for each possible position, scale, and rotation. 

Assuming that the different part evidence maps are conditionally independent, and that each $\mathbf{d}_i$ depends only on the part configuration $\mathbf{l}_i$, the likelihood $p(D|L)$ can be written as:
\begin{equation}
\label{eq:PSlikelihood}
p(D|L) = \prod_{i=0}^{N} p\left(\mathbf{d}_i|\mathbf{l}_i\right).
\end{equation}

Substituting Eq.~\eqref{eq:PSprior} and Eq.~\eqref{eq:PSlikelihood} in Eq.~\eqref{eq:PSposterior}, one finally obtains:
\begin{equation}
p(L|D) \propto p(\mathbf{l}_0) \cdot \prod_{i=0}^{N} p\left(\mathbf{d}_i|\mathbf{l}_i\right) \cdot \prod_{\substack{(i,j)\in E}} p\left(\mathbf{l}_i|\mathbf{l}_j\right)
\end{equation}

The part detectors $p\left(\mathbf{d}_i|\mathbf{l}_i\right)$ use a variant of the shape context descriptor \cite{Mikolajczyk2005}, that consists in a log-polar histogram of locally normalised gradient orientations. The feature vector is obtained by concatenating all shape context descriptors whose centres fall inside the bounding box of the part.
During detection, different positions, scales, and orientations are scanned with sliding windows. 
The classifier used for detection is an ensemble of a fixed number of decision stumps combined through AdaBoost \cite{Freund1995}.

\subsection{Features}
\label{sec:featuresSOA}
Each body part (or the whole image of the individual, if no body part subdivision model is used) is typically described using one or more different global or local features. In this, Section, the main kinds of features used in the literature are reviewed.

\subsubsection{Global features}
Global features are characteristics measured in the whole image or body region considered, and are usually represented as a fixed-size vector of real numbers.
 
Probably the most widely used feature of this kind is the global colour histogram. Given a colour image of size $N=W\times H$ pixels, the colours of the image are at first quantised into $B$ bins $1,\ldots,B$. The histogram is then constructed as the count of the number of occurrences per bin. Typically, such count is normalised as the fraction of pixels of the image belonging to the bin.
Colour image pixels are typically represented as a triplet of values, representing the amount of colour in different colour channels (e.g., Red, Green and Blue). In this case, each colour channel is quantised separately. The resulting histogram can be multi-dimensional (one dimension for each channel), or mono-dimensional (the final histogram is constructed as the concatenation of histograms in each colour channel). The latter saves a lot of space (e.g., if 16 bin are used for each colour channel, the size of the multi-dimensional histogram would be $16*16*16=4096$ bins, while the mono-dimensional one would have a size of 48 bins) and has usually a similar discriminant capability to the former.
Various colour spaces exist in the literature. Among them it is worth citing:
\begin{itemize}
\item The RGB colour space, where each colour is represented as the corresponding amount of Red, Green and Blue; it directly relates to the way devices acquire and visualise colours.
\item \emph{Perceptual} colour spaces, i.e., spaces inspired to the way the human brain perceives colour; e.g., the Hue-Saturation-Value (HSV) colour space, in which the light intensity (V channel) is separated from the colour tonality (H channel) and the saturation of the colour (S channel).
\end{itemize}
Good surveys on colour spaces are provided in \cite{Tkalcic2003,vandeSande2010}.
Many appearance descriptors use global colour histograms, to represent the whole body appearance \cite{Bazzani2010,Javed2005,Liu2012} or the overall appearance of each body part \cite{Avraham2012,Bazzani2012,Bedagkar2012,Bedagkar2011,Farenzena2010,Gheissari2006,Gray2008ELF,Khan2010,ProsserReIdBMVC2010, Wu2011,Wei-Shi2011}. 
Du et al. \cite{Du2012} evaluated the use of colour histograms computed in various colour spaces for building appearance descriptors for re-identification.
To tackle with the lower amount of information usually carried by peripheral pixels (that could actually belong to the background, as the person segmentation is usually very noisy), in \cite{Cheng2011,Farenzena2010,Wu2011} these pixels receive less weight than those near the vertical silhouette symmetry axis.

The colour space is typically quantised in an uniform fashion. However, many colour ranges can be irrelevant for representing a certain appearance, e.g. colours ranges that are not present in the image, or whose coverage percentage with respect to the image is irrelevant. For this reason, some approaches try first to find the most representative colour ranges, then describe the appearance with respect to these ones. One of the methods of \cite{Bak2010-2} and the methods of \cite{Bedagkar2012,Bedagkar2011,Kai2011} use the Dominant Colour Descriptor (DCD) (also called Representative Meta Colours Model, RMCM) of MPEG-7, which provides a compact description of the most representative colours. Given an image, the DCD algorithm first finds the $K$ \emph{dominant colours} \cite{Deng01}, via \emph{k-means} clustering of all the colour triplets in the image. Then, the descriptor is defined as
\begin{equation}
F = \big\{\{c_i,p_i\}, i = 1,\ldots,K \big\}
\end{equation}
where $c_i$ is the $i$-th dominant colour (i.e., the centroid of the $i$-th cluster), and $p_i$ is the percentage of image pixels that fall into the $i$-th cluster. A similar approach is used also in \cite{Cai2010}, called Global Colour Context.
The method of \cite{Dangelo2011} partly differs to the former ones, although it shares with them the same idea of describing appearance in terms of the most important colours. Instead of finding representative colours by clustering, they are chosen a priori; specifically, eleven colors, usually referred to as \emph{culture colours} \cite{Dangelo2010}, are used: black, white, red, yellow, green, blue, brown, purple, pink, orange, and grey. Each pixel of the image is assigned to the most similar cultural colour.

Colour histograms are invariant to scale and show a good robustness with respect to partial occlusions, if the occlusion itself is small.
However, they are sensitive to changing brightness and colour response of the sensor. Illumination conditions in outdoor environments may consistently vary during time due to changing weather conditions and the varying illumination of the Sun during the day. On the other hand, lighting conditions of indoor scenes may vary from camera to camera due to different types of lamps (e.g., incandescent, tungsten, neon) and also due to weather conditions in case of presence of windows that let the Sun light enter.
Colour response of the sensors may also vary due to environmental conditions and due to the automatic colour balance that often takes place in-camera. 

Different mechanisms have been exploited to address, at least partially, the above problems.
Probably the simplest one is colour normalisation \cite{vandeSande2010}. 
The chromaticity RGB space is one of these techniques, used in \cite{Bouma2012,Du2012,TruongCong2010}, and consists of dividing each colour channel of each pixel by the sum of all the channels of that pixel, e.g. $R'=R/(R+G+B)$.
Another common technique is the \emph{Grey-world normalisation} \cite{Buchsbaum1980}, which relies on the assumption that the average colour of a scene is usually a tonality of grey. It consists of dividing each RGB channel of every pixel by the average value of that channel in the image, e.g. $R'=R / \operatorname*{mean}(R)$. Grey-world normalisation is used in \cite{Truong2009,TruongCong2010}.
Similar to Grey-world is the affine normalisation used in \cite{Bouma2012,Truong2009,TruongCong2010}, where pixel-values of each color channel are normalised independently by subtracting the average and scaling them with the standard deviation, e.g. $R'=\big(R -\operatorname*{mean}(R)\big) / \operatorname*{std}(R)$.

Alternative to colour normalisation is histogram equalisation \cite{Finlayson2005}, which is used in the re-identification methods of \cite{Bak2010,Truong2009,TruongCong2010}. It is based on the assumption that a change in illumination preserves the rank ordering of sensor responses (i.e. pixel values). The rank measure for the $i-th$ bin of the histogram and the $k$-th colour channel is defined as $M_k(i)=\sum_{u=0}^{i}H_k(u)/\sum_{u=0}^{N}H_k(u)$, where $N$ is the number of bins and $H_k()$ is the histogram relative to the $k$-th channel.

Finally, Piccardi and Cheng \cite{Piccardi2005} exploited a colour quantisation scheme to mitigate the effect of illumination changes between cameras. They represent the image with a Major Colour Spectrum Histogram (MCSH), that is, an histogram of the top $N$ represented colour values in the image.

Another problem of histograms is that they do not retain any information on the spatial disposition of colours. 
A simple way to incorporate the spatial information is to add the relative pixel height (i.e. the ratio between the vertical coordinate of the pixel and the total height of the silhouette) as another channel of the image\footnote{The horizontal coordinate of the pixel is typically not used, as it is not robust to body rotations and viewpoint changes.}. A colour-position histogram can be then built which is able to spatially localise the colour distribution \cite{Bouma2012,Truong2009,TruongCong2010}.
A similar approach is used also in \cite{Khan2010}, where two dimensions are added to each pixel (i.e. the radial and angular distance to the torso center) and quantised.
The Color Structure Descriptor (CSD) of MPEG-7 \cite{Manjunath2001} is used in \cite{Hahnel2004}, and encodes the distribution of 
colour by the following steps: (i) move a window of size $8\times 8$ pixel over the picture ; (ii) determine which colours are present in within the window; (iii) increase the corresponding bins in a color histogram  by one, independently of the number of pixels of these colors. 

Instead of looking at colour properties, other kinds of global features try to characterise gradients, textures and repeated patterns of the whole body appearance or of each body part. Gabor filters \cite{Movellan2008} ans Schmid filters \cite{Schmid01cvpr} are orientation-sensitive filters that capture texture and edge informations on the image. The former ones are aimed at detecting horizontal and vertical lines, while the latter ones detect circular gradient changes.
They are used in various appearance descriptors \cite{Gray2008ELF,Liu2012,Ma2012,ProsserReIdBMVC2010,Wei-Shi2011} in conjunction with other colour-related features.

Hahnel et al. \cite{Hahnel2004} compared various different texture features. The fist is the 2D Quadrature Mirror Filter (QMF), a well known filter in signal processing that splits a 2D input signal into two bands (high and low-pass) in each direction (horizontal, vertical and diagonal. The second is the Oriented Gaussian Derivatives (OGD) filter, based on steerable Gaussian filters. 
Also, two MPEG-7 texture-related descriptors, are used the Homogeneous Texture Descriptor (HTD) that uses Gabor filters, and the Edge Histogram Descriptor (EHD), basically an histograms of the directions of each edge pixel in the image \cite{Sikora2001}.

It is worth pointing out that texture-based features have always been used in combination to colour-based ones. Information on repeated patterns is in fact likely to be not distinctive enough when used alone. Hahnel et al. \cite{Hahnel2004} confirmed this thought, and showed also that the combination of colour and texture-based descriptors may lead only to minor performance improvements.

\subsubsection{Local features}
The term \emph{local feature} refers to an appearance characteristic of a small portion of the image (e.g., the neighbourhood of a pixel). The regions where local features are extracted can be chosen in various way (e.g. by dense sampling, by an interest operator or at random). Each small region is described by a feature vector (e.g., an histogram).
This lead to a representation of the image as as a \emph{bag} (set) of local features.

\emph{Interest points} are one important category of local features. The most famous among them is SIFT (Scale Invariant Feature Transform) \cite{Lowe2004}, where at first salient points of the image are chosen via in interest operator that looks for ``stable'' locations in the image (i.e. locations that are identifiable over different scales and rotations). This operation is carried out by detecting scale-extrema locations in the \emph{scale space} of scale $\sigma$, which is defined by the function
\begin{equation}
\label{eq:ScaleSpace1}
L(x,y,\sigma) = \mathcal{N}(x,y,\sigma) \ast I(x,y)
\end{equation}
where $\ast$ is the convolution operation in the image coordinates $x$ and $y$, and $\mathcal{N}(x,y,\sigma)$ is a 2-D Gaussian with standard deviation $\sigma$. Stable key-points can be detected in this space e.g. by using difference-of-Gaussians functions convolved with the image:
\begin{equation}
\label{eq:ScaleSpace2}
\begin{aligned}
D(x,y,\sigma) = \big(\mathcal{N}(x,y,k\sigma) - \mathcal{N}(x,y,k\sigma)\big) \ast I(x,y) = \\
L(x,y,k\sigma) - L(x,y,\sigma)
\end{aligned}
\end{equation}

To detect the local minima and maxima of $D(x,y,\sigma)$, each point $(x,y)$ is compared with its 8 neighbours at the same scale $k\sigma$, and its 9 neighbours in the two scales $(k-1)\sigma$ and ($k+1\sigma$). If this value is the minimum or maximum of all these points, then this point is an extrema, and it is labelled as key-point.
A subsequent stage filters out low-contrast and noisy points. The remaining key-points are described as a histogram of the edge orientations of a small window centred on the key-point.
SIFT points or its variants, (e.g., Speeded-Up Robust Features, SURF \cite{Bay2008}) are used in various appearance descriptors to represent the whole body appearance. Interest point are typically chosen via interest operators \cite{Oliveira2009,Hamdoun2008,Hamdoun2008b,Kai2011,Ma2012,Martinel2012EL,Martinel2012} but some works exist (e.g.,\cite{zhao2013}) that adopt dense sampling instead.

Other approaches use different kinds of local features.

\begin{figure}[t]
\centering
   \includegraphics[width=1.00\linewidth]{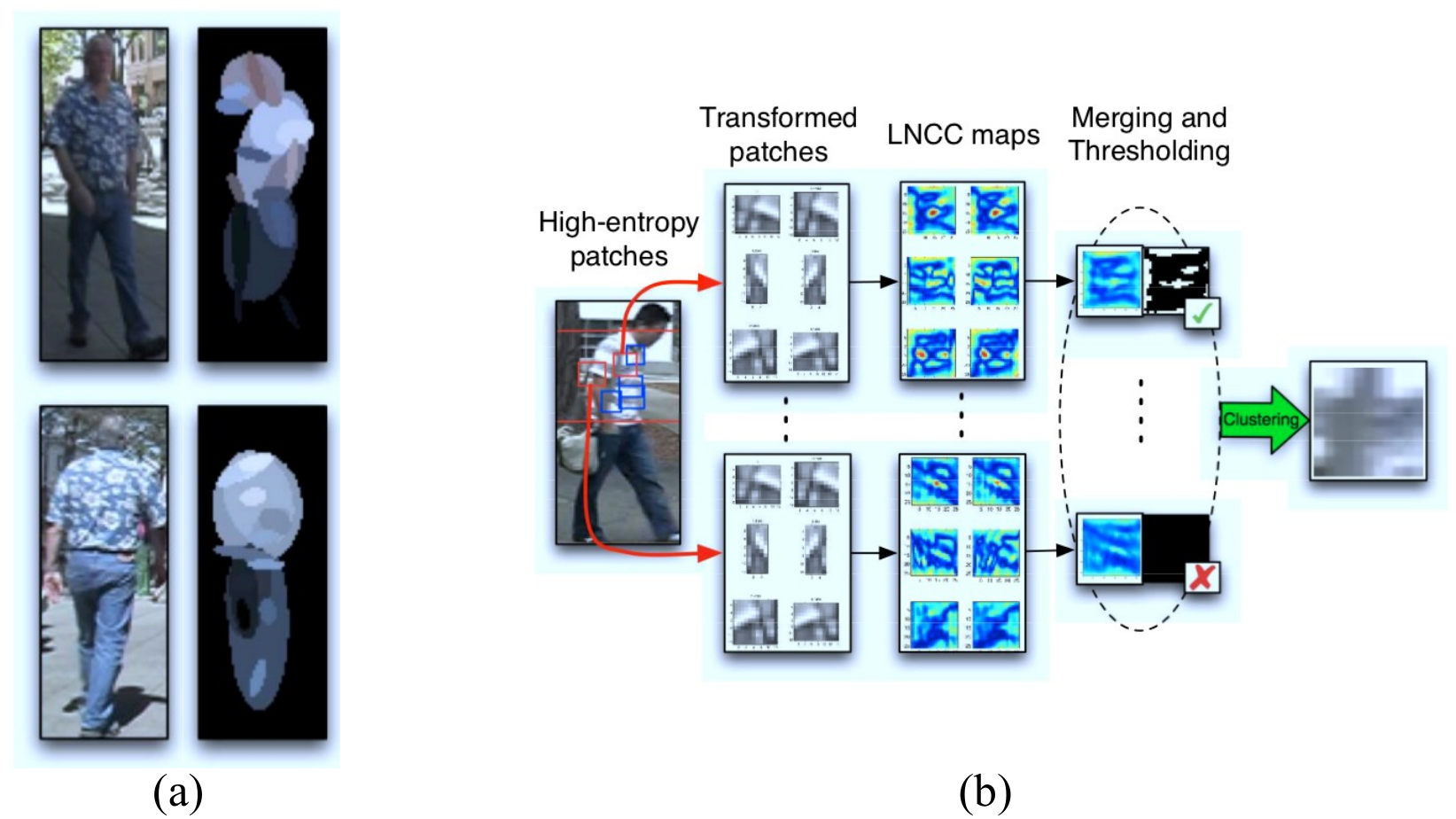}
   \caption{(a) Maximally Stable Colour Regions \cite{f07b} detected in two images showing the same pedestrian. (b) Steps of the extraction of RHSP: random extraction, rotational invariance check, recurrence check, entropy thresholding, clustering. The final result of this process is a set of patches (in this case only one) characterising repeated patterns of each body part of the individual. Figures taken from \cite{Farenzena2010}.}
\label{fig:localFeatures1}
\end{figure}

Maximally Stable Colour Regions (MSCR) \cite{f07b} are used in \cite{Cheng2011,Farenzena2010,Ma2012}. The MSCR algorithm first detects a set of regions in the image (Fig.~\ref{fig:localFeatures1}-a) by using a constrained agglomerative clustering on image pixels, which show the maximal chromatic distance. The detected regions are then described by their area, centroid, second moment matrix and average color, forming 9-dimensional feature vectors, and are stable to scale and affine transforms.

Recurrent Highly-Structured Patches (RHSP) used in the method of \cite{Farenzena2010}, try instead to capture repeated patterns and textures of the clothing appearance. The procedure of creating RHSPs is as follows.
First, random and possibly overlapping small patches are extracted from the image. Patches that do not carry texture informations (e.g. showing uniform colours) are discarded by thresholding the patch entropy, computed as the sum of the entropy of each colour channel.
Remaining patches are then further filtered, keeping only those that exhibit invariance to rotations.
Second, the recurrence of each patch is evaluated, via Local Normalised Cross-Correlation over a small local region containing that patch. Third, patches that show a high degree of recurrence are clustered, maintaining for each final cluster the patch nearest to the centroid. These patches are finally described as their Local Binary Pattern histogram \cite{Ojala02}, a simple yet efficient way to describe textured content, based on a per-pixel transform that encodes small-scale appearance structures. 

Instead of using interest operators like the one defined by Eqs.\eqref{eq:ScaleSpace1}-\eqref{eq:ScaleSpace2}, or other proper selection criteria to choose where to extract a local feature, in \cite{Gray2008ELF}, a set of strips of fixed height and position are extracted from the image, and described by a concatenation of colour histograms in different colour spaces and Gabor and Shmid filters. Similarly, in \cite{Hirzer2012} partly overlapping rectangular patches of fixed size are sampled from the image following a pre-defined regular grid. Each patch is represented by its colour histogram in the HSV colour space, and by its LBP histogram to capture textures and repeated patterns. An analogous approach is also used in \cite{zhao2013}, except for the fact that patches are not overlapping. 
Finally, instead of using regular sampling, one could sample patches at random, an approach followed for instance in \cite{Satta2011MCM}.

To reduce the dimensionality of local features-based descriptors, in \cite{Satta2011MCD,Satta2012MCD} a \textit{dissimilarity} approach has been introduced \cite{Pekalska2005}: a bag of local features is turned into a dissimilarity-vector that encodes the degree of similarity to a set of predefined prototype local features. Prototypes are found by clustering local features extracted from a design set of images of people. In case a part-based body model is used, memberships to body parts are kept and each body part is represented via a dedicated dissimilarity vector.
The same dissimilarity-based descriptor was then used in \cite{Satta2012MMSEC,Satta2012peopleSearch,Satta2013Kinect}, also for tasks different that person re-identification.
 
\subsection{Combination of features and matching}
\label{sec:combinationSOA}
Many person re-identification methods use appearance descriptors made up of only one kind of features among the above mentioned ones, typically based on colour or interest points \cite{Avraham2012,Bak2010-2,Bedagkar2012,Bedagkar2011,Bouma2012,Cai2010,Dangelo2011,Gheissari2006,Hamdoun2008,Hamdoun2008b,Khan2010,
Martinel2012EL,Martinel2012,Truong2009,TruongCong2010}.
However, as combining different sources of information usually helps in attaining a better performance, especially when sources are complementary (i.e. they look at different aspects of the appearance, e.g. colour and texture), many authors have defined descriptors that use a combination of features.

In principle, two main combination techniques can be exploited to this aim \cite{Ross2004}:\footnote{In \emph{verification} tasks, whose goal is to establish whether the claimed identity is true, combination can also be performed at \emph{decision} level, i.e., by combining the crisp outputs of classifier/detectors. It can not be applied to person re-identification, which is a \emph{recognition} task instead.}
\begin{enumerate}
\item \emph{feature}-level fusion: if the features used are made up of a single vector of fixed size (e.g. global features, or local features with an intrinsic ordering) they can be combined simply by concatenating feature vectors;
\item \emph{score}-level fusion: a distinct detector/matcher is used for each feature, and their real-valued scores are combined (e.g., by averaging them, or using their maximum value).
\end{enumerate}
The first approach is followed for instance in \cite{Du2012,Hirzer2012,Wu2011}. 
The second approach requires to define a proper fusion rule.
Many methods used a weighted average of the partial scores attained with each single feature, where weights are fixed a-priori by the system designer \cite{Bazzani2010,Bazzani2012,Cheng2011,Farenzena2010}.
Another approach is to learn a proper metric or a set of weights from a training set.
In \cite{Gray2008ELF}, AdaBoost\cite{Freund1995} is used to this aim:  each feature set is associated to a weak two-class classifier (a decision stump) which discerns between the class 0 (identities differ) and 1 (identity is the same) based only in that feature set. The method of \cite{ProsserReIdBMVC2010} tries to find a linear function to weight the absolute difference of samples by training an ensemble of RankSVM rankers \cite{Joachims2002} given pairwise relevance constraints.
The Probabilistic Relative Distance Comparison (PRDC) technique of \cite{Wei-Shi2011} maximises the probability that a pair of true match has a smaller distance than that of a wrong match. The output is an orthogonal matrix which essentially encodes the global importance of each feature.
In \cite{Ma2012} a pairwise metric is learned through a recently proposed method, Pairwise Constrained Component Analysis (PCCA) \cite{Mignon2012}, which learns a projection into a low-dimensional space where the distance between pairs of data points respects the desired constraints.

Metric learning and similar approaches always help in boosting re-identification performance. However, it is worth to note that all the above methods require a training set of labelled data. Such set can be for instance the gallery of templates. This requires that the template gallery is \emph{fixed}, i.e. templates cannot be added during system operation; such constraint might be too strong for real-world application scenarios.

\section{Other cues}
\label{sec:othercuesSOA}
Some cues alternative to the clothing appearance have been exploited in the literature to perform person re-identification or assimilable tasks. Despite the intrinsic limitations of such cues, they could be potentially of help in certain conditions, possibly combined with appearance cues.

\begin{figure}[t]
\centering
   \includegraphics[width=0.99\linewidth]{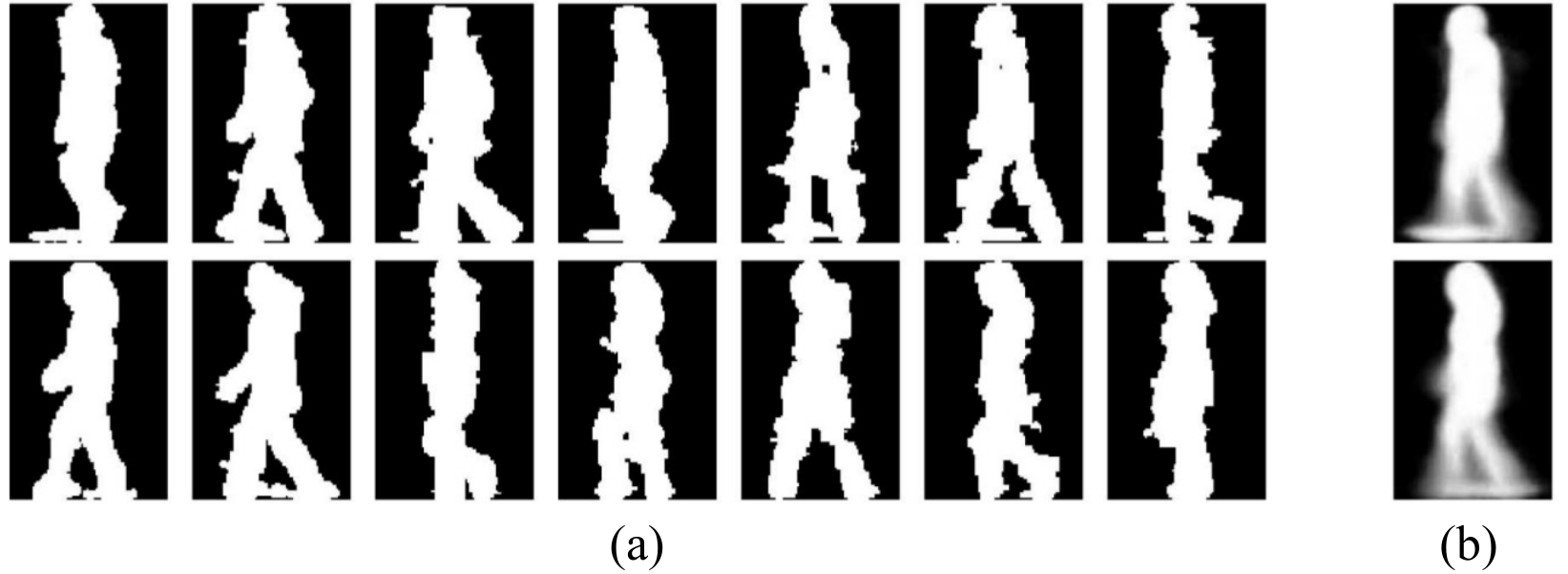}
   \caption{(a) Two sequences of aligned foreground silhouettes. (b) Their corresponding Gait Energy Image. Figures taken from \cite{Han2006}.}
\label{fig:gei}
\end{figure}

Human \emph{gait}, i.e. the recurrent pattern of motion of a person walking, is among these cues. In cognitive science, it is known to be one of the cues that humans exploit to recognise people \cite{Stevenage1999}. Among the approaches to characterise gait, the recently proposed Gait Energy Image (GEI) \cite{Han2006} has attracted the attention of many researchers. Here, the gait signature is formed by by normalising, aligning and averaging a sequence of foreground silhouettes corresponding to one ``walking period'' (see Fig.~\ref{fig:gei}).
Principal Component Analysis (PCA) is then used to reduce the dimensionality of the signature.

The use of Gait Energy Image can lead to high recognition rates \cite{Xu2012} and can overcome one of the main limitations of clothing appearance-based approaches, that is, the impossibility of distinguishing people when their clothing changes between observations. It is also not directly affected by illumination changes. However, it requires perfect alignments of the silhouettes to be compared, and is sensible to segmentation errors. These two constraints severely limit the use of GEI-based methods on practical, real-world applications.
Researchers have therefore attempted to explore other approaches. Zhao et al. \cite{Zhao2006} and more recently Gu et al. \cite{Gu2010} used a 3D skeletal representation, that however requires multiple overlapping camera views or a constrained environment to construct and track it.

Some authors attempted instead to perform \emph{remote face recognition} \cite{Ni2010}, that is, face recognition with low resolution images. As low resolution face images are not directly usable for recognition, many approaches attempted to address the problem through the obvious way of trying to increase image resolution, using super-resolution techniques \cite{Gunturk2003,Hennings2008,Jia2005,Shekhar2011}.
Other authors proposed instead techniques that work directly on low resolution images, by exploiting metric learning \cite{Li2010,Li2009},  multidimensional scaling \cite{Biswas2012}, or multiple frames from video sequences \cite{Arandjelovic2006}.
All the approaches above could in principle be used in conjunction with appearance cues to increase re-identification accuracy when the face is visible. 

Another useful set of soft cues is anthropometry, that is, the characterisation of individuals through the measurement of physical body features \cite{Roebuck1975}, e.g., height, arm length, and eye-to-eye distance. Measures are typically taken according to a number of body landmark points (e.g., elbows, hands, knees, feet), that have to be localized either automatically or manually.
In the classic study by Daniels and Churchill \cite{Daniels1952}, the uniqueness of 10 different anthropometric traits was evaluated on a large data base of 4063 individuals.
None of the considered traits was found to be ``average'' (i.e., approximately close to the mean point), considering all 10 dimensions. Furthermore, only 7\% of the individuals were ``average'' in 2 dimensions, and 3\% in 3 dimensions.

Although the use of anthropometric measurements for person recognition has been proposed in many works, their extraction was often based on costly devices, like 3D laser scanners, and/or require user collaboration in a constrained environment \cite{Godil2003, Neugebauer2009,Ober2010}.
In some works, anthropometric measurements are extracted from a single RGB camera view, instead.
In \cite{Barron2001} a method that does not require camera calibration was proposed, for simultaneously estimating anthropometric measurements and pose. However, the former are measured up to a scale factor, and consequently can not be used to directly compare individuals in images acquired by different cameras.
Calibration is not required in \cite{BenAbdelkader2008} as well, although 13 body landmarks have to be manually selected, from an image of an individual in frontal pose.
Other methods focus on height measurement only \cite{BenAbdelkader2008Height,Gallagher2009,Lee2012,Lee2010,Madden2005}, but require camera calibration to estimate absolute height values.
Interestingly, in \cite{Madden2005} height is used as a cue for the task of associating tracks of individuals coming from disjoint camera views, which is actually the same \emph{re-acquisition} task that is enabled by person re-identification.

\begin{figure}[t]
\centering
   \includegraphics[width=0.99\linewidth]{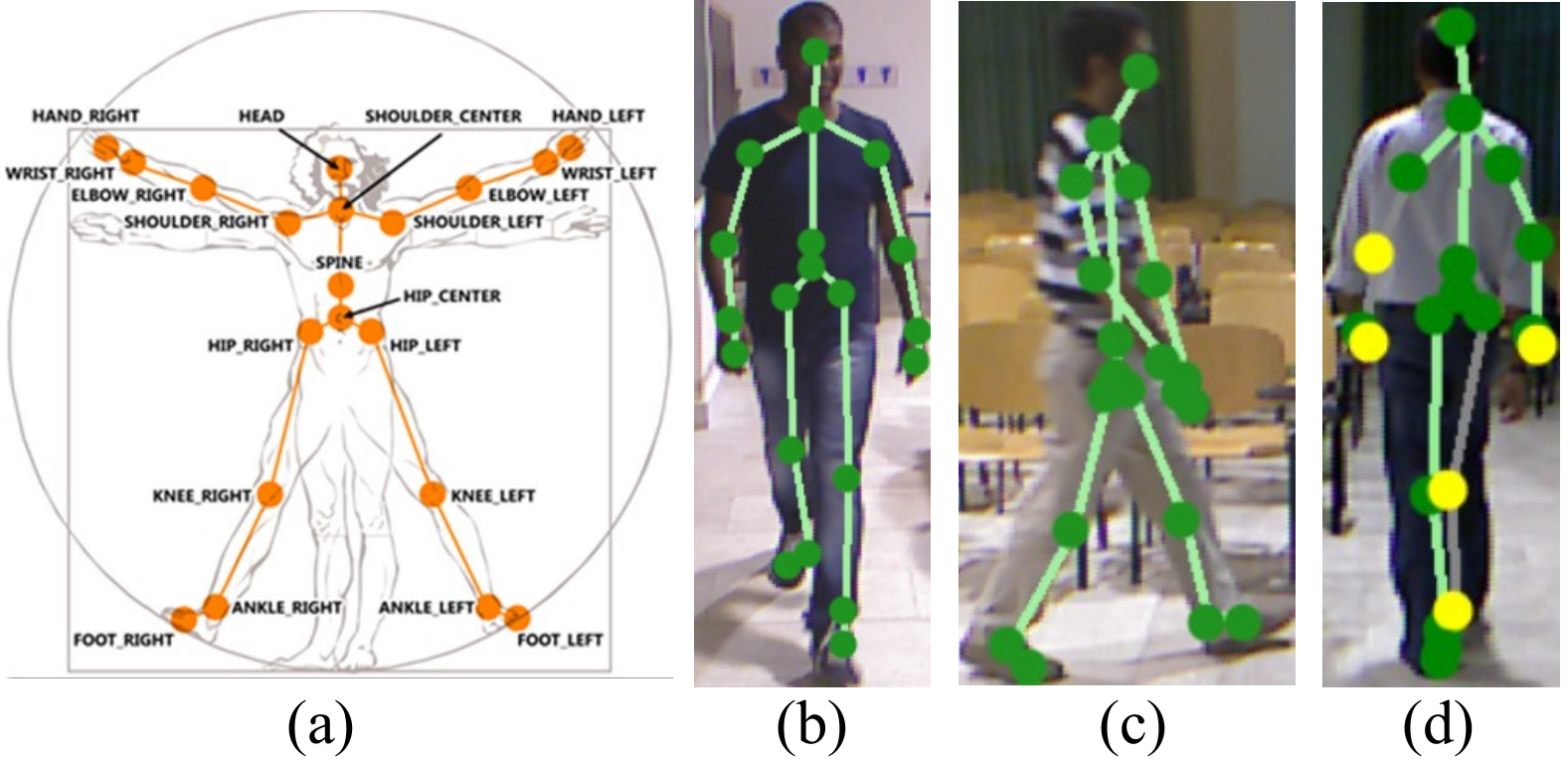}
\caption{(a) The 20 skeletal points tracked by the Kinect SDK in the classical representation of the Vitruvian Man. (b--d) Examples of the pose estimation capabilities of the Kinect SDK. Depending on the degree of confidence of the estimation of the points position, the Kinect SDK distinguishes between \emph{good} (in green) or \emph{inferred} (in yellow) points, the latter being less reliable than the former.}
\label{fig:VitruvianMan}
\end{figure}

None of the above works fits the typical setting of person re-identification tasks, which is characterised by multiple, uncalibrated cameras and unconstrained environment, with free poses and non collaborative users.
Recently, it has been shown that body pose can be reliably estimated in real-time by exploiting RGB-D sensors \cite{Shotton2011, Taylor2012}, like the MS Kinect, a device recently introduced in the video-gaming market.
The pose estimation functionality of Kinect SDK \cite{KinectSDK}, which is based on a similar method, provides the absolute position (in meters) of 20 different body joints in real-time, with high reliability (see Fig.~\ref{fig:VitruvianMan}).
Detecting joint positions enables the evaluation of several anthropometric measures.
In \cite{Barbosa2012} such joints were used to extract a set of different anthropometric measures from front or back poses: distance between floor and head, ratio between torso and legs, height, distance between floor and neck, distance between neck and left shoulder, distance between neck and right shoulder, and distance between torso center and right shoulder.
Other three geodesic distance measures were estimated from the 3D mesh of the abdomen, obtained from the Kinect depth map:
torso center to left shoulder, torso center (located in the abdomen) to left hip, and between torso center to right hip.
Results reported in \cite{Barbosa2012} appear promising. However, many of the considered anthropometric measures are hard or impossible to extract from unconstrained poses. For instance, extracting measures from 3D mesh requires near-frontal pose (abdomen is hidden in back pose); neck distance to left and right shoulders becomes hard to compute from lateral pose, even using a depth map, and requires to distinguish between left and right body parts. Such issues limit the actual set of anthropometric measures that can be used in realistic scenarios.

\section{Conclusions}
\label{sec:conclusions}
This paper provided a survey of current approaches and methods for constructing appearance descriptors for person re-identification. State-of-the-art descriptors have been reviewed from two different viewpoints, namely the kind of body model and the kind of features used to represent a person.
We tried to provide a comprehensive analysis and description of the algorithms in a structured and consolidated way. We hope that this work will be a useful reference for anyone in the research community willing to work on this interesting and challenging topic. 

\bibliographystyle{plain}

\begin{thebibliography}{100}

\bibitem{BenAbdelkader2008}
C.~Ben Abdelkader and Y.~Yacoob.
\newblock Statistical estimation of human anthropometry from a single
  uncalibrated image.
\newblock {\em Computational Forensics}, 2008.

\bibitem{Amit1996}
Yali Amit and Augustine Kong.
\newblock Graphical templates for model registration.
\newblock {\em IEEE Transactions on Pattern Analysis and Machine Intelligence},
  18(3):225--236, mar 1996.

\bibitem{Andriluka2008tracking}
M.~Andriluka, S.~Roth, and B.~Schiele.
\newblock People-tracking-by-detection and people-detection-by-tracking.
\newblock In {\em Proceedings of the IEEE Conference on Computer Vision and
  Pattern Recognition (CVPR)}, pages 1--8, june 2008.

\bibitem{Andriluka2009Pictorials}
M.~Andriluka, S.~Roth, and B.~Schiele.
\newblock Pictorial structures revisited: People detection and articulated pose
  estimation.
\newblock In {\em Proceedings of the 2009 IEEE Conference on Computer Vision
  and Pattern Recognition (CVPR)}, pages 1014--1021, 2009.

\bibitem{Arandjelovic2006}
Ognjen Arandjelovic and Roberto Cipolla.
\newblock Face recognition from video using the generic shape-illumination
  manifold.
\newblock In {\em Proceedings of the 9th European Conference on Computer Vision
  (ECCV)}, pages 27--40, 2006.

\bibitem{Avraham2012}
Tamar Avraham, Ilya Gurvich, Michael Lindenbaum, and Shaul Markovitch.
\newblock Learning implicit transfer for person re-identification.
\newblock In {\em Proceedings of the European Conference of Computer Vision
  (ECCV) Workshops, 1st Workshop on Re-Identification (REID)}, pages 381--390,
  2012.

\bibitem{Ayedi2012}
Walid Ayedi, Hichem Snoussi, and Mohamed Abid.
\newblock A fast multi-scale covariance descriptor for object
  re-identification.
\newblock {\em Pattern Recognition Letters}, 33(14):1902--1907, 2012.
\newblock Special Issue on Novel Pattern Recognition-Based Methods for
  Re-identification in Biometric Context.

\bibitem{Bak2010-2}
Slawomir Bak, Etienne Corvee, Francois Bremond, and Monique Thonnat.
\newblock Person re-identification using haar-based and dcd-based signature.
\newblock In {\em Proceedings of the 7th IEEE International Conference on
  Advanced Video and Signal Based Surveillance (AVSS)}, pages 1--8, 2010.

\bibitem{Bak2010}
Slawomir Bak, Etienne Corvee, Francois Bremond, and Monique Thonnat.
\newblock Person re-identification using spatial covariance regions of human
  body parts.
\newblock In {\em Proceedings of the 7th IEEE International Conference on
  Advanced Video and Signal Based Surveillance (AVSS)}, pages 435--440, 2010.

\bibitem{Barbosa2012}
B.~I. Barbosa, M.~Cristani, A.~Del~Bue, L.~Bazzani, and V.~Murino.
\newblock Re-identification with rgb-d sensors.
\newblock In {\em Proceedings of the European Conference of Computer Vision
  (ECCV) Workshops, 1st Workshop on Re-Identification (REID)}, 2012.

\bibitem{Barron2001}
Carlos Barr\'{o}n and Ioannis~A. Kakadiaris.
\newblock Estimating anthropometry and pose from a single uncalibrated image.
\newblock {\em Computer Vision and Image Understanding}, 81(3):269--284, March
  2001.

\bibitem{Bay2008}
Herbert Bay, Andreas Ess, Tinne Tuytelaars, and Luc Van~Gool.
\newblock Speeded-up robust features (surf).
\newblock {\em Computuer Vision and Image Understanding}, 110(3):346--359, June
  2008.

\bibitem{Bazzani2010}
Loris Bazzani, Marco Cristani, Alessandro Perina, Michela Farenzena, and
  Vittorio Murino.
\newblock Multiple-shot person re-identification by hpe signature.
\newblock In {\em Proceedings of the 20th International Conference on Pattern
  Recognition (ICPR)}, pages 1413--1416, Washington, DC, USA, 2010. IEEE
  Computer Society.

\bibitem{Bazzani2012}
Loris Bazzani, Marco Cristani, Alessandro Perina, and Vittorio Murino.
\newblock Multiple-shot person re-identification by chromatic and epitomic
  analyses.
\newblock {\em Pattern Recognition Letters}, 33(7):898--903, 2012.
\newblock Special Issue on Awards from ICPR 2010.

\bibitem{Bedagkar2012}
A.~Bedagkar-Gala and Shishir~K. Shah.
\newblock Part-based spatio-temporal model for multi-person re-identification.
\newblock {\em Pattern Recognition Letters}, 33(14):1908--1915, October 2012.

\bibitem{Bedagkar2011}
Apurva Bedagkar-Gala and Shishir~K. Shah.
\newblock Multiple person re-identification using part based spatio-temporal
  color appearance model.
\newblock In {\em Proceedings of the 2011 IEEE International Conference on
  Computer Vision Workshops (ICCV Workshops)}, pages 1721--1728, nov. 2011.

\bibitem{BenAbdelkader2008Height}
C.~BenAbdelkader and Y.~Yacoob.
\newblock Statistical body height estimation from a single image.
\newblock In {\em Proceedings of the 8th IEEE International Conference on
  Automatic Face Gesture Recognition (FG)}, pages 1--7, 2008.

\bibitem{Biswas2012}
Soma Biswas, Kevin~W. Bowyer, and Patrick~J. Flynn.
\newblock Multidimensional scaling for matching low-resolution face images.
\newblock {\em IEEE Transactions on Pattern Analysis and Machine Intelligence},
  34(10):2019--2030, 2012.

\bibitem{Bouma2012}
Henri Bouma, Sander Borsboom, Richard J.~M. den Hollander, Sander~H. Landsmeer,
  and Marcel Worring.
\newblock Re-identification of persons in multi-camera surveillance under
  varying viewpoints and illumination.
\newblock {\em Proceedings SPIE 8359, Sensors, and Command, Control,
  Communications, and Intelligence (C3I) Technologies for Homeland Security and
  Homeland Defense XI}, pages 83590Q--83590Q--10, 2012.

\bibitem{Bouwmans2008}
Thierry Bouwmans, Fida El~Baf, and Bertrand Vachon.
\newblock {B}ackground {M}odeling using {M}ixture of {G}aussians for
  {F}oreground {D}etection - {A} {S}urvey.
\newblock {\em Recent Patents on Computer Science}, 1(3):219--237, November
  2008.

\bibitem{Buchsbaum1980}
G.~Buchsbaum.
\newblock A spatial processor model for object colour perception.
\newblock {\em Journal of the Franklin Institute}, 310(1):1--26, 1980.

\bibitem{Burges1998}
Christopher J.~C. Burges.
\newblock A tutorial on support vector machines for pattern recognition.
\newblock {\em Data Mining and Knowledge Discovovery}, 2(2):121--167, June
  1998.

\bibitem{Cai2010}
Yinghao Cai and Matti Pietik\"{a}inen.
\newblock Person re-identification based on global color context.
\newblock In {\em Proceedings of the Tenth International Workshop on Visual
  Surveillance (VS)}, ACCV'10, pages 205--215, Berlin, Heidelberg, 2011.
  Springer-Verlag.

\bibitem{Canny1986}
John Canny.
\newblock A computational approach to edge detection.
\newblock {\em IEEE Transactions on Pattern Analysis and Machine Intelligence},
  8(6):679--698, nov. 1986.

\bibitem{Cheng2011}
Dong~S. Cheng, Marco Cristani, Michele Stoppa, Loris Bazzani, and Vittorio
  Murino.
\newblock Custom pictorial structures for re-identification.
\newblock In {\em Proceedings of the British Machine Vision Conference (BMVC)},
  pages 68.1--68.11, 2011.

\bibitem{Dangelo2010}
{A}ngela {D}'angelo and {J}ean-{L}uc {D}ugelay.
\newblock A statistical approach to culture colors distribution in video
  sensors.
\newblock In {\em 5th International Workshop on Video Processing and Quality
  Metrics for Consumer Electronics (VPQM)}, {S}cottsdale, AZ, {United}
  {States}, 01 2010.

\bibitem{Dangelo2011}
{A}ngela {D}'angelo and {J}ean-{L}uc {D}ugelay.
\newblock {P}eople re-identification in camera networks based on probabilistic
  color histograms.
\newblock In {\em {SPIE} 2011, {E}lectronic {I}maging {C}onference on 3{D}
  {I}mage {P}rocessing (3{DIP}) and {A}pplications, {V}ol. 7882, 23-27
  {J}anuary, 2011, {S}an {F}rancisco, {CA}, {USA}}, {S}an {F}rancisco, {United}
  {States}, 01 2011.

\bibitem{Daniels1952}
G.~S. Daniels and E.~Churchill.
\newblock The average man?
\newblock {\em Technical Note WCRD TN 53-7: Wright-Patterson Air Force Base,
  OH: Wright Air Force Development Center}, 1952.

\bibitem{Oliveira2009}
I.O. de~Oliveira and J.L. de~Souza~Pio.
\newblock Object reidentification in multiple cameras system.
\newblock In {\em Proceedings of the 4th International Conference on Embedded
  and Multimedia Computing (EM-Com)}, pages 1--8, 2009.

\bibitem{Deng01}
Yining Deng, B.~S. Manjunath, Charles Kenney, Michael~S. Moore, Student Member,
  and Hyundoo Shin.
\newblock An efficient color representation for image retrieval.
\newblock {\em IEEE Transactions on Image Processing}, 10:140--147, 2001.

\bibitem{Dollar2012}
P.~Dollar, C.~Wojek, B.~Schiele, and P.~Perona.
\newblock Pedestrian detection: An evaluation of the state of the art.
\newblock {\em IEEE Transactions on Pattern Analysis and Machine Intelligence},
  34(4):743--761, april 2012.

\bibitem{Doretto2011}
Gianfranco Doretto, Thomas Sebastian, Peter Tu, and Jens Rittscher.
\newblock Appearance-based person reidentification in camera networks: problem
  overview and current approaches.
\newblock {\em Journal of Ambient Intelligence and Humanized Computing},
  2:127--151, 2011.

\bibitem{Du2012}
Yuning Du, Haizhou Ai, and Shihong Lao.
\newblock Evaluation of color spaces for person re-identification.
\newblock In {\em Proceedings of the 21st International Conference on Pattern
  Recognition (ICPR)}, Washington, DC, USA, 2012. IEEE Computer Society.

\bibitem{DudaHart2001}
Richard~O. Duda, Peter~E. Hart, and David~G. Stork.
\newblock {\em Pattern Classification}.
\newblock Wiley, New York, 2. edition, 2001.

\bibitem{Elgammal2000}
Ahmed~M. Elgammal, David Harwood, and Larry~S. Davis.
\newblock Non-parametric model for background subtraction.
\newblock In {\em Proceedings of the 6th European Conference on Computer
  Vision-Part II}, ECCV '00, pages 751--767, London, UK, UK, 2000.
  Springer-Verlag.

\bibitem{Farenzena2010}
Michela Farenzena, Loris Bazzani, Alessandro Perina, Vittorio Murino, and Marco
  Cristani.
\newblock Person re-identification by symmetry-driven accumulation of local
  features.
\newblock In {\em Proceedings of the 2010 IEEE Conference on Computer Vision
  and Pattern Recognition (CVPR)}, pages 2360--2367, 2010.

\bibitem{Felzenszwalb08}
Pedro Felzenszwalb, David McAllester, and Deva Ramanan.
\newblock A discriminatively trained, multiscale, deformable part model.
\newblock In {\em Proceedings of the 2008 IEEE Conference on Computer Vision
  and Pattern Recognition (CVPR)}, 2008.

\bibitem{Felzenszwalb2005}
Pedro~F. Felzenszwalb.
\newblock Representation and detection of deformable shapes.
\newblock {\em IEEE Transactions on Pattern Analysis and Machine Intelligence},
  27(2):208--220, feb. 2005.

\bibitem{Felzenszwalb2005PS}
Pedro~F. Felzenszwalb and Daniel~P. Huttenlocher.
\newblock Pictorial structures for object recognition.
\newblock {\em International Journal of Computer Vision}, 61(1):55--79, January
  2005.

\bibitem{Finlayson2005}
Graham Finlayson, Steven Hordley, Gerald Schaefer, and Gui~Yun Tian.
\newblock Illuminant and device invariant colour using histogram equalisation.
\newblock {\em Pattern Recognition}, 38(2):179--190, 2005.

\bibitem{f07b}
Per-Erik Forss\'en.
\newblock Maximally stable colour regions for recognition and matching.
\newblock In {\em Proceedings of the IEEE Conference on Computer Vision and
  Pattern Recognition (CVPR)}, Minneapolis, USA, June 2007. {IEEE} Computer
  Society.

\bibitem{Freund1995}
Yoav Freund and Robert~E. Schapire.
\newblock A decision-theoretic generalization of on-line learning and an
  application to boosting.
\newblock In {\em Proceedings of the Second European Conference on
  Computational Learning Theory}, EuroCOLT '95, pages 23--37, London, UK, UK,
  1995. Springer-Verlag.

\bibitem{Gallagher2009}
Andrew~C. Gallagher, Andrew~C. Blose, and Tsuhan Chen.
\newblock Jointly estimating demographics and height with a calibrated camera.
\newblock In {\em Proceedings of the IEEE 12th International Conference on
  Computer Vision (ICCV)}, pages 1187--1194, 2009.

\bibitem{Gheissari2006}
Niloofar Gheissari, Thomas~B. Sebastian, and Richard Hartley.
\newblock Person reidentification using spatiotemporal appearance.
\newblock In {\em Proceedings of the 2006 IEEE Conference on Computer Vision
  and Pattern Recognition (CVPR)}, volume~2, pages 1528--1535, 2006.

\bibitem{Godil2003}
Afzal Godil, Patrick Grother, and Sandy Ressler.
\newblock Human identification from body shape.
\newblock In {\em Proceedings of the 4th International Conference on 3D Digital
  Imaging and Modeling (3DIM)}, pages 386--393, 2003.

\bibitem{Gray2007evaluating}
Douglas Gray, S.~Brennan, and H.~Tao.
\newblock Evaluating appearance models for recognition, reacquisition, and
  tracking.
\newblock In {\em Proceedings of the 10th IEEE International Workshop on
  Performance Evaluation of Tracking and Surveillance (PETS)}, pages 41--47,
  2007.

\bibitem{Gray2008ELF}
Douglas Gray and Hai Tao.
\newblock Viewpoint invariant pedestrian recognition with an ensemble of
  localized features.
\newblock In {\em Proceedings of the 10th European Conference on Computer
  Vision (ECCV)}, pages 262--275, 2008.

\bibitem{Gu2010}
Junxia Gu, Xiaoqing Ding, Shengjin Wang, and Youshou Wu.
\newblock Action and gait recognition from recovered 3-d human joints.
\newblock {\em Transaction on System, Man and Cybernetics, Part B},
  40(4):1021--1033, August 2010.

\bibitem{Gunturk2003}
Bahadir~K. Gunturk, Aziz~Umit Batur, Yucel Altunbasak, Monson H.~Hayes III, and
  Russell~M. Mersereau.
\newblock Eigenface-domain super-resolution for face recognition.
\newblock {\em IEEE Transactions on Image Processing}, 12(5):597--606, 2003.

\bibitem{Hahnel2004}
M.~Hahnel, D.~Klunder, and K.-F. Kraiss.
\newblock Color and texture features for person recognition.
\newblock In {\em Proceedings of the 2004 IEEE International Joint Conference
  on Neural Networks}, volume~1, july 2004.

\bibitem{Hamdoun2008}
O.~Hamdoun, F.~Moutarde, B.~Stanciulescu, and B.~Steux.
\newblock Interest points harvesting in video sequences for efficient person
  identification.
\newblock In {\em Proceedings of the 8th International Workshop on Visual
  Surveillance (VS)}, 2008.

\bibitem{Hamdoun2008b}
O.~Hamdoun, F.~Moutarde, B.~Stanciulescu, and B.~Steux.
\newblock Person re-identification in multi-camera system by signature based on
  interest point descriptors collected on short video sequences.
\newblock In {\em Proceedings of the Second ACM/IEEE International Conference
  on Distributed Smart Cameras, 2008. ICDSC 2008.}, pages 1--6, sept. 2008.

\bibitem{Han2006}
Ju~Han and Bir Bhanu.
\newblock Individual recognition using gait energy image.
\newblock {\em IEEE Transactions on Pattern Analisys and Machine Intelligence},
  28(2):316--322, February 2006.

\bibitem{Hennings2008}
Pablo~H. Hennings-Yeomans, Simon Baker, and B.~V. K.~Vijaya Kumar.
\newblock Simultaneous super-resolution and feature extraction for recognition
  of low-resolution faces.
\newblock In {\em Proceedings of the 2008 IEEE Conference on Computer Vision
  and Pattern Recognition (CVPR)}. IEEE Computer Society, 2008.

\bibitem{Hirzer2012}
Martin Hirzer, Peter~M. Roth, and Horst Bischof.
\newblock Person re-identification by efficient impostor-based metric learning.
\newblock In {\em Proceedings of the Ninth IEEE International Conference on
  Advanced Video and Signal-Based Surveillance, (AVSS)}, pages 203--208, 2012.

\bibitem{Isard2001}
M.~Isard and J.~MacCormick.
\newblock Bramble: a bayesian multiple-blob tracker.
\newblock In {\em Proceedings of the Eighth IEEE International Conference on
  Computer Vision (ICCV)}, volume~2, pages 34--41, 2001.

\bibitem{Javed2005}
Omar Javed, Khurram Shafique, and Mubarak Shah.
\newblock Appearance modeling for tracking in multiple non-overlapping cameras.
\newblock In {\em Proceedings of the 2005 IEEE Conference on Computer Vision
  and Pattern Recognition (CVPR)}, volume~2, pages 26--33, june 2005.

\bibitem{Jia2005}
Kui Jia and Shaogang Gong.
\newblock Multi-modal tensor face for simultaneous super-resolution and
  recognition.
\newblock In {\em Proceedings of the Tenth IEEE International Conference on
  Computer Vision - Volume 2}, ICCV '05, pages 1683--1690, Washington, DC, USA,
  2005. IEEE Computer Society.

\bibitem{Joachims2002}
Thorsten Joachims.
\newblock Optimizing search engines using clickthrough data.
\newblock In {\em Proceedings of the eighth ACM SIGKDD international conference
  on Knowledge discovery and data mining}, KDD '02, pages 133--142, New York,
  NY, USA, 2002. ACM.

\bibitem{Kai2011}
Kai Jungling, C.~Bodensteiner, and M.~Arens.
\newblock Person re-identification in multi-camera networks.
\newblock In {\em Proceedings of the 2011 IEEE Conference on Computer Vision
  and Pattern Recognition Workshops (CVPRW)}, pages 55--61, june 2011.

\bibitem{Khan2010}
Arif Khan, Jian Zhang, and Yang Wang.
\newblock Appearance-based re-identification of people in video.
\newblock In {\em Proceedings of the 2010 International Conference on Digital
  Image Computing: Techniques and Applications (DICTA)}, pages 357--362, dec.
  2010.

\bibitem{KinectSDK}
Microsoft\textregistered Kinect\texttrademark.
\newblock http://www.microsoft.com/en-us/kinectforwindows/.

\bibitem{Lee2012}
Kual-Zheng Lee.
\newblock A simple calibration approach to single view height estimation.
\newblock In {\em Proceedings of the 9th Conference on Computer and Robot
  Vision}, pages 161--166, 2012.

\bibitem{Lee2010}
Seok-Han Lee, Tae-Eun Kim, and Jong-Soo Choi.
\newblock A single-view based framework for robust estimation of heights and
  positions of moving people.
\newblock In {\em Digest of Technical Papers of the 2010 International
  Conference on Consumer Electronics (ICCE)}, pages 503--504, 2010.

\bibitem{Li2010}
B.~Li, H.~Chang, S.~Shan, and X.~Chen.
\newblock Low-resolution face recognition via coupled locality preserving
  mappings.
\newblock {\em IEEE Signal Processing Letters}, 17(1):20--23, January 2010.

\bibitem{Li2009}
Bo~Li, Hong Chang, Shiguang Shan, and Xilin Chen.
\newblock Coupled metric learning for face recognition with degraded images.
\newblock In {\em Proceedings of the 1st Asian Conference on Machine Learning:
  Advances in Machine Learning}, ACML '09, pages 220--233, Berlin, Heidelberg,
  2009. Springer-Verlag.

\bibitem{Liu2012}
Chunxiao Liu, Shaogang Gong, Chen~Change Loy, and Xinggang Lin.
\newblock Person re-identification: What features are important?
\newblock In {\em Proceedings of the European Conference of Computer Vision
  (ECCV) Workshops, 1st Workshop on Re-Identification (REID)}, 2012.

\bibitem{Lowe2004}
David~G. Lowe.
\newblock Distinctive image features from scale-invariant keypoints.
\newblock {\em International Journal of Computer Vision}, 60(2):91--110,
  November 2004.

\bibitem{Ma2012}
Bipeng Ma, Yu~Su, and Frederic Jurie.
\newblock Local descriptors encoded by fisher vectors for person
  re-identification.
\newblock In {\em Proceedings of the European Conference of Computer Vision
  (ECCV) Workshops, 1st Workshop on Re-Identification (REID)}, pages 413--422,
  2012.

\bibitem{Madden2005}
C.~Madden and M.~Piccardi.
\newblock Height measurement as a session-based biometric for people matching
  across disjoint camera views.
\newblock In {\em In Image and Vision Computing New Zealand}, page~29, 2005.

\bibitem{Manjunath2001}
B.S. Manjunath, J.-R. Ohm, V.V. Vasudevan, and A.~Yamada.
\newblock Color and texture descriptors.
\newblock {\em IEEE Transactions on Circuits and Systems for Video Technology},
  11(6):703--715, jun 2001.

\bibitem{Martinel2012EL}
Niki Martinel and Gian~Luca Foresti.
\newblock Multi-signature based person re-identification.
\newblock {\em Electronics Letters}, 48(13):765--767, 21 2012.

\bibitem{Martinel2012}
Niki Martinel and Christian Micheloni.
\newblock Re-identify people in wide area camera network.
\newblock In {\em Proceedings of the 2012 IEEE Conference on Computer Vision
  and Pattern Recognition Workshops (CVPRW)}, pages 31--36, june 2012.

\bibitem{Mignon2012}
Alexis Mignon.
\newblock Pcca: A new approach for distance learning from sparse pairwise
  constraints.
\newblock In {\em Proceedings of the 2012 IEEE Conference on Computer Vision
  and Pattern Recognition (CVPR)}, CVPR '12, pages 2666--2672, Washington, DC,
  USA, 2012. IEEE Computer Society.

\bibitem{Mikolajczyk2005}
Krystian Mikolajczyk and Cordelia Schmid.
\newblock A performance evaluation of local descriptors.
\newblock {\em IEEE Transaction on Pattern Analysis and Maching Intelligence},
  27(10):1615--1630, October 2005.

\bibitem{Movellan2008}
Javier~R. Movellan.
\newblock {Tutorial on Gabor Filters}.
\newblock {\em Tutorial paper http://mplab.ucsd.edu/tutorials/pdfs/gabor.pdf},
  2008.

\bibitem{Neugebauer2009}
S.~P. Neugebauer and P.~A. Sallee.
\newblock New 3d biometric capabilities for human identification at a distance.
\newblock In {\em Proceedings of the 2009 Special Operations Forces Industry
  Conference (SOFIC)}, 2009.

\bibitem{Ni2010}
Jie Ni and Rama Chellappa.
\newblock Evaluation of state-of-the-art algorithms for remote face
  recognition.
\newblock In {\em Proceedings of the 2010 International Conference on Image
  Processing (ICIP)}, pages 1581--1584, 2010.

\bibitem{Niu2003}
W.~Niu, L.~Jiao, D.~Han, and Y.-F. Wang.
\newblock Real-time multiperson tracking in video surveillance.
\newblock In {\em Proceedings of the 2003 Joint Conference of the Fourth
  International Conference on Information, Communications and Signal
  Processing, and Fourth Pacific Rim Conference on Multimedia}, volume~2, pages
  1144--1148, dec. 2003.

\bibitem{Ober2010}
D.B. Ober, S.P. Neugebauer, and P.A. Sallee.
\newblock Training and feature-reduction techniques for human identification
  using anthropometry.
\newblock In {\em Proceedings of the Fourth IEEE International Conference on
  Biometrics: Theory Applications and Systems (BTAS)}, pages 1 --8, Sept. 2010.

\bibitem{Ojala02}
Timo Ojala, Matti Pietikainen, and Topi Maenpaa.
\newblock Multiresolution gray-scale and rotation invariant texture
  classification with local binary patterns.
\newblock {\em IEEE Transactions on Pattern Analysis and Machine Intelligence},
  24(7):971--987, 2002.

\bibitem{Pekalska2005}
Elzbieta Pekalska and Robert P.~W. Duin.
\newblock {\em The Dissimilarity Representation for Pattern Recognition:
  Foundations And Applications}, volume~64 of {\em Machine Perception and
  Artificial Intelligence}.
\newblock World Scientific Publishing Co., Inc., River Edge, NJ, USA, 2005.

\bibitem{Piccardi2005}
Massimo Piccardi and Eric~Dahai Cheng.
\newblock Track matching over disjoint camera views based on an incremental
  major color spectrum histogram.
\newblock In {\em Proceedings of the 2005 IEEE International Conference on
  Video and Signal Based Surveillance (AVSS 05), 15-16 September 2005, Como,
  Italy}, pages 147--152. IEEE Computer Society, 2005.

\bibitem{ProsserReIdBMVC2010}
B.~Prosser, W.~Zheng, S.~Gong, and T.~Xiang.
\newblock Person re-identification by support vector ranking.
\newblock In {\em Proceedings of the British Machine Vision Conference (BMVC)},
  pages 21.1--21.10, 2010.

\bibitem{Ramanan2006}
Deva Ramanan.
\newblock Learning to parse images of articulated bodies.
\newblock In {\em Proceedings of the Conference on Neural Information
  Processing Systems (NIPS)}, 2006.

\bibitem{Roebuck1975}
J.A. Roebuck, K.H.E. Kroemer, and W.G. Thomson.
\newblock {\em Engineering anthropometry methods}.
\newblock Wiley series in human factors. Wiley-Interscience, 1975.

\bibitem{Ross2004}
A.~Ross and A.~K. Jain.
\newblock {Multimodal Biometrics: an overview}.
\newblock In {\em Proceedings of 12th European Signal Processing Conference},
  pages 1221--1224, 2004.

\bibitem{Satta2013PHD}
Riccardo Satta.
\newblock {\em Dissimilarity-based people re-identification and search for
  intelligent video surveillance}.
\newblock Phd thesis, PhD School in Information Engineering, University of
  Cagliari, Cagliari (Italy), 2013.

\bibitem{Satta2011MCD}
Riccardo Satta, Giorgio Fumera, and Fabio Roli.
\newblock Exploiting dissimilarity representations for person
  re-identification.
\newblock In {\em Proceedings of the 1st International Workshop on
  Similarity-Based Pattern Analysis and Recognition (SIMBAD)}, pages 275--289,
  2011.

\bibitem{Satta2012MMSEC}
Riccardo Satta, Giorgio Fumera, and Fabio Roli.
\newblock Appearance-based people recognition by local dissimilarity
  representations.
\newblock In {\em Proceedings of the ACM Workshop on Multimedia and Security},
  MMSec '12, pages 151--156, New York, NY, USA, 2012. ACM.

\bibitem{Satta2012MCD}
Riccardo Satta, Giorgio Fumera, and Fabio Roli.
\newblock Fast person re-identification based on dissimilarity representations.
\newblock {\em Pattern Recognition Letters}, 33(14):1838--1848, 2012.

\bibitem{Satta2012peopleSearch}
Riccardo Satta, Giorgio Fumera, and Fabio Roli.
\newblock A general method for appearance-based people search based on textual
  queries.
\newblock In {\em Proceedings of the European Conference of Computer Vision
  (ECCV) Workshops, 1st Workshop on Re-Identification (REID)}, volume 7583,
  pages 453--461. Springer Berlin Heidelberg, 2012.

\bibitem{Satta2011MCM}
Riccardo Satta, Giorgio Fumera, Fabio Roli, Marco Cristani, and Vittorio
  Murino.
\newblock A multiple component matching framework for person re-identification.
\newblock In {\em Proceedings of the 16th International Conference on Image
  Analysis and Processing (ICIAP)}, volume~2, pages 140--149, 2011.

\bibitem{Satta2013Kinect}
Riccardo Satta, Federico Pala, Giorgio Fumera, and Fabio Roli.
\newblock Real-time appearance-based person re-identification over multiple
  kinecttm cameras.
\newblock In {\em Proceedings of the 8th International Conference on Computer
  Vision Theory and Applications (VISAPP 2013)}, Barcelona, Spain, 21/02/2013
  2013.

\bibitem{Schmid01cvpr}
Cordelia Schmid.
\newblock Constructing models for content-based image retrieval.
\newblock In {\em Proceedings of the IEEE Conference on Computer Vision and
  Pattern Recognition (CVPR)}, volume~2, pages 39--45, 2001.

\bibitem{Shekhar2011}
Sumit Shekhar, Vishal~M. Patel, and Rama Chellappa.
\newblock Synthesis-based recognition of low resolution faces.
\newblock In {\em Proceedings of the 2011 International Joint Conference on
  Biometrics}, IJCB '11, pages 1--6, Washington, DC, USA, 2011. IEEE Computer
  Society.

\bibitem{Shotton2011}
J.~Shotton, A.~Fitzgibbon, M.~Cook, T.~Sharp, M.~Finocchio, R.~Moore,
  A.~Kipman, and A.~Blake.
\newblock Real-time human pose recognition in parts from single depth images.
\newblock In {\em Proceedings of the 2011 IEEE Conference on Computer Vision
  and Pattern Recognition (CVPR)}, pages 1297--1304, 2011.

\bibitem{Sigal06}
Leonid Sigal and Michael~J. Black.
\newblock Predicting 3d people from 2d pictures.
\newblock In {\em Proceedings of the IV Conference on Articulated Motion and
  Deformable Objects (AMDO)}, pages 185--195, 2006.

\bibitem{Sikora2001}
Thomas Sikora.
\newblock The mpeg-7 visual standard for content description - an overview.
\newblock {\em IEEE Transactions on Circuits and Systems for Video Technology},
  11(6):696--702, June 2001.

\bibitem{Stevenage1999}
Sarah~V. Stevenage, Mark~S. Nixon, and Kate Vince.
\newblock Visual analysis of gait as a cue to identity.
\newblock {\em Applied Cognitive Psychology}, 13(6):513--526, 1999.

\bibitem{Taylor2012}
J.~Taylor, J.~Shotton, T.~Sharp, and A.~Fitzgibbon.
\newblock The vitruvian manifold: Inferring dense correspondences for one-shot
  human pose estimation.
\newblock In {\em Proceedings of the 2012 IEEE Conference on Computer Vision
  and Pattern Recognition (CVPR)}, pages 103--110, 2012.

\bibitem{Tkalcic2003}
M.~Tkalcic and J.F. Tasic.
\newblock Colour spaces: perceptual, historical and applicational background.
\newblock In {\em EUROCON 2003. Computer as a Tool. The IEEE Region 8},
  volume~1, pages 304--308, sept. 2003.

\bibitem{Truong2009}
Dung~Nghi Truong~Cong, Catherine Achard, Louahdi Khoudour, and Lounis Douadi.
\newblock Video sequences association for people re-identification across
  multiple non-overlapping cameras.
\newblock In {\em Proceedings of the 15th International Conference on Image
  Analysis and Processing (ICIAP)}, pages 179--189, 2009.

\bibitem{TruongCong2010}
Dung~Nghi Truong~Cong, Louahdi Khoudour, Catherine Achard, Cyril Meurie, and
  Olivier Lezoray.
\newblock People re-identification by spectral classification of silhouettes.
\newblock {\em Signal Processing}, 90(8):2362--2374, August 2010.

\bibitem{vandeSande2010}
Koen van~de Sande, Theo Gevers, and Cees Snoek.
\newblock Evaluating color descriptors for object and scene recognition.
\newblock {\em IEEE Transactions on Pattern Analysis and Machine Intelligence},
  32(9):1582--1596, September 2010.

\bibitem{Wu2011}
Yang Wu, Masayuki Mukunoki, Takuya Funatomi, M.~Minoh, and Shihong Lao.
\newblock Optimizing mean reciprocal rank for person re-identification.
\newblock In {\em Proceedings of the 2011 8th IEEE International Conference on
  Advanced Video and Signal Based Surveillance}, AVSS '11, pages 408--413,
  Washington, DC, USA, 2011. IEEE Computer Society.

\bibitem{Xu2012}
Dong Xu, Yi~Huang, Zinan Zeng, and Xinxing Xu.
\newblock Human gait recognition using patch distribution feature and
  locality-constrained group sparse representation.
\newblock {\em IEEE Transactions on Image Processing}, 21(1):316--326, jan.
  2012.

\bibitem{Yang2008}
Nai-Chung Yang, Wei-Han Chang, Chung-Ming Kuo, and Tsia-Hsing Li.
\newblock A fast mpeg-7 dominant color extraction with new similarity measure
  for image retrieval.
\newblock {\em Journal of Visual Communication and Image Representation},
  19(2):92--105, February 2008.

\bibitem{Yilmaz2006}
Alper Yilmaz, Omar Javed, and Mubarak Shah.
\newblock Object tracking: A survey.
\newblock {\em ACM Computing Surveys (CSUR)}, 38(4), December 2006.

\bibitem{Zhao2006}
Guoying Zhao, Guoyi Liu, Hua Li, and M.~Pietikainen.
\newblock 3d gait recognition using multiple cameras.
\newblock In {\em 7th International Conference on Automatic Face and Gesture
  Recognition (FGR)}, pages 529--534, april 2006.

\bibitem{zhao2013}
Rui Zhao, Wanli Ouyang, and Xiaogang Wang.
\newblock Unsupervised salience learning for person re-identification.
\newblock In {\em Proceedings of the IEEE Conference on Computer Vision and
  Pattern Recognition (CVPR)}, Portland, USA, June 2013.

\bibitem{Zheng2009ilids}
Wei-Shi Zheng, Shaogang Gong, and Tao Xiang.
\newblock Associating groups of people.
\newblock In {\em Proceedings of the British Machine Vision Conference (BMVC)},
  2009.

\bibitem{Wei-Shi2011}
Wei-Shi Zheng, Shaogang Gong, and Tao Xiang.
\newblock Person re-identification by probabilistic relative distance
  comparison.
\newblock In {\em Proceedings of the 2011 IEEE Conference on Computer Vision
  and Pattern Recognition (CVPR)}, pages 649--656, Washington, DC, USA, 2011.
  IEEE Computer Society.

\end{thebibliography}

\end{document}